\documentclass[10pt,twocolumn,letterpaper]{article}

\usepackage{cvpr}              

\usepackage[dvipsnames]{xcolor}
\usepackage{xfrac}
\usepackage{wrapfig}
\usepackage{multicol}
\usepackage{graphicx}
\usepackage[accsupp]{axessibility}  
\newcommand{\norm}[1]{\left \lVert #1 \right \rVert}

\definecolor{cvprblue}{rgb}{0.21,0.49,0.74}
\usepackage[pagebackref,breaklinks,colorlinks,citecolor=cvprblue]{hyperref}

\def\paperID{7929} 
\def\confName{CVPR}
\def\confYear{2024}

\title{Building Optimal Neural Architectures using Interpretable Knowledge}

\author{
Keith G. Mills$^{1, 2}$\quad  
Fred X. Han$^2$\quad  
Mohammad Salameh$^2$\quad  
Shengyao Lu$^1$\\
Chunhua Zhou$^3$\quad  
Jiao He$^3$\quad  
Fengyu Sun$^3$\quad  
Di Niu$^1$\\
$^1$Dept. ECE, University of Alberta\quad $^2$Huawei Technologies Canada\quad $^3$Huawei Kirin Solution, China\\
{\tt\small \{kgmills, shengyao, dniu\}@ualberta.ca\quad sunfengyu@hisilicon.com}\\ 
{\tt\small \{fred.xuefei.han1, mohammad.salameh, zhouchunhua, hejiao4\}@huawei.com}\\ 
}

\begin{document}
\maketitle
\begin{abstract}

Neural Architecture Search is a costly practice. The fact that a search space can span a vast number of design choices with each architecture evaluation taking nontrivial overhead makes it hard for an algorithm to sufficiently explore candidate networks. In this paper, we propose AutoBuild, a scheme which learns to align the latent embeddings of operations and architecture modules with the ground-truth performance of the architectures they appear in. By doing so, AutoBuild is capable of assigning interpretable importance scores to architecture modules, such as individual operation features and larger macro operation sequences such that high-performance neural networks can be constructed without any need for search. Through experiments performed on state-of-the-art image classification, segmentation, and Stable Diffusion models, we show that by mining a relatively small set of evaluated architectures, AutoBuild can learn to build high-quality architectures directly or help to reduce search space to focus on relevant areas, finding better architectures that outperform both the original labeled ones and ones found by search baselines. Code available at \url{https://github.com/Ascend-Research/AutoBuild}

\end{abstract}
\section{Introduction}
\label{sec:intro}

Neural Architecture Search (NAS) is an AutoML technique for neural network design and has been widely adopted in computer vision. Given a task, most NAS methods involve exploring a vast search space of candidate architectures,    
e.g.,~\cite{pham2018ENAS, mehta2022suite, wan2020fbnetv2, Dai2021FBNetV3JA}, using a search algorithm, to find the best architecture in terms of predefined performance and hardware-friendly metrics.  
However, NAS is a costly procedure mainly due to two reasons. First, the search space is formed by varying the operation configurations in each layer and is thus exponential to the number of configurations per layer in scale, which can easily exceed $10^{10}$ architectures~\cite{liu2018DARTS, cai2020once}, making it hard to thoroughly explore the search space. 
Second, there is a large cost associated with each architecture evaluation, which requires full training before assessing performance on a validation set.

Several methods have been proposed in the literature to reduce the evaluation cost in NAS. For example, Once-for-All (OFA)~\cite{cai2020once} trains a weight-sharing supernetwork that can represent any architecture in the search space. 
Recently, SnapFusion~\cite{li2023snapfusion}, which is adopted by Qualcomm's latest Snapdragon 8 Gen 3 Mobile Platform, applies a robust training procedure to Stable Diffusion models, so that the U-Net is trained to be robust to architectural permutations and block pruning, which enables it to find a pruned efficient U-Net that speeds up text-to-image generation significantly. In fact, these methods still pay an initial, very high one-time computational cost to train the supernetworks or robust U-Net (which requires a tremendous investment into GPUs), such that during search individual candidate architectures can be assessed by pulling weights from the supernetwork. 
Another limitation is that supernetwork training is specific to the dataset used, e.g., ImageNet~\cite{deng2009imagenet}, and retraining may incur a significant cost. As foundational models~\cite{li2023snapfusion, Rombach_2022_CVPR, openai2023gpt4}  grow in complexity, NAS is becoming infeasible for organizations with limited computing resources.

In this paper, we propose AutoBuild, a method for automatically constructing optimal neural architectures from high-valued architecture building blocks discovered from only a handful of random architectures that have been evaluated.
Unlike traditional NAS that considers how to best traverse a vast search space, AutoBuild focuses on discovering the features, operations and multi-layered subgraphs as well as their configurations--in general, interpretable knowledge--that are important to performance in order to directly construct architectures. The intuition is that although the size of the search space is exponential large, the actual number of configurations per layer can be small. We ask the question--could there be a method that learns how to build a good architecture rather than searching for one,
thus circumventing the formidable search cost?

Specifically, we learn the importance of architecture modules using a ranking loss mechanism applied to each hop-level of a Graph Neural Network (GNN)~\cite{brody2022how, velivckovic2017graph, xu2019GIN} predictor to align the learnt embeddings for each subgraph or node with the ground-truth predictor targets (e.g., accuracy). This ensures important subgraphs and nodes will have higher embedding norms (by the GNN) if they appear in architectures with higher performance. We propose methods to be able to rank operation-level features, e.g., convolution kernel size vs. input/output latent tensor size for a CNN, or rank operations, e.g., attention vs. residual convolutions in a diffusion model, or rank subgraphs of operation sequences. AutoBuild can then construct high-quality architectures by combining high valued modules in each layer or can greatly prune the size of original search space prior to performing NAS. Moreover, we can readily manipulate predictor labels to combine multiple performance metrics, e.g., both accuracy and latency, to focus on different regions of the search space.

Through extensive experiments performed on state-of-the-art image classification, segmentation, and Stable Diffusion models, we show that by mining a relatively small set of evaluated architectures, AutoBuild can learn to assign importance `scores' to the building blocks such that a high-quality architecture can be directly constructed without search or searched for in a much reduced search space to generate better architectures.

To prove the effectiveness of AutoBuild, we show that by learning from 3000  labeled architectures for ImageNet, AutoBuild can focus on shortlisted regions of the vast search space of MobileNetV3 to generate superior Pareto frontier of accuracy vs. latency compared to other schemes that search the whole search space or reduced search space based on manual insights. A similar trend holds for panoptic segmentation~\cite{kirillov2019panoptic}. Furthermore, in a search space consisting of variants of Stable Diffusion v1.4~\cite{Rombach_2022_CVPR} by tweaking the configurations of each layer of its U-Net, where we have access to only 68 labeled architectures, AutoBuild can quickly generate a architecture without search that achieves better image Inpainting performance as compared to the labeled architectures or a predictor-based exhaustive search strategy. 
\section{Related Work}
\label{sec:related}

Search Spaces fall into two categories: Macro-level and micro-level. Macro-level spaces represent architectures as a sequence of predefined structures, e.g., MBConv blocks in MobileNets~\cite{howard2017mobilenets, sandler2018mobilenetv2, howard2019searching, wan2020fbnetv2, Dai2021FBNetV3JA}, or even Attention vs. Convolution layers in Stable Diffusion U-Nets~\cite{Rombach_2022_CVPR, podell2023sdxl}. Representing the network as a sequence implicitly encodes some level of spatial information, such as latent tensor channel size and stride, which can change across the depth of the network. By contrast, micro-level spaces, like many NAS-Benchmarks~\cite{mehta2022suite}, designs the search space over a cell structure that is repeated multiple times to form the entire network. The cell takes the form of different DAGs over a predefined list of operations and typically do not encode spatial information. 
Although many of the techniques in AutoBuild apply to both macro and micro-search spaces, we focus on macro-level spaces like MobileNets and Stable Diffusion U-Nets in this paper. 

Neural performance predictors~\cite{white2021powerful, lu2023pinat, mills2023gennape} learn to estimate the performance of candidate architectures and overcome the computational burden of training them from scratch, in order to facilitate more lightweight NAS. By contrast, AutoBuild focuses on learning in a way that quantifies the importance of architecture modules. AutoBuild further leverages these insights to reduce the overall search space size, and thus, the necessity to perform NAS.

Interpretable approaches generate insights on the relationship between search spaces and performance metrics. For example, NAS-BOWL~\cite{Ru2021InterpretableNA} uses Weisfeiler-Lehman (WL) kernels to extract architecture motifs. However, NAS-BOWL mainly operates on micro search spaces as the WL-kernel does not efficiently scale with graph depth~\cite{salameh2024autogo}. Sampling-based approaches~\cite{mills2021profiling} examine how architecture modules impact accuracy and latency. However, these methods rely on large amounts of data to generate statistical information. In contrast, AutoBuild is more data-driven and efficient - capable of learning insights from limited data.

Ranking losses have previously been used to structure GNN embedding spaces. CT-NAS~\cite{chen2021contrastive} use a ranking loss to determine which architecture in a pair has higher performance. AutoBuild uses a Magnitude Ranked Embedding Space to generate importance scores, which is partially inspired by the Ordered Embedding Space that SPMiner~\cite{ying2024representation} construct to perform frequent subgraph mining.
\section{Background}
\label{sec:background}

This section provides background information on how architectures can be cast as graphs. Further, we elaborate on how node embeddings produced by GNNs are used to `score' their corresponding rooted subgraphs. 

\subsection{Architecture Graph Encodings}
\label{sec:arch_formats}

A common way to represent candidate architectures for NAS is as a Directed Acyclic Graph (DAG). A DAG is a graph $\mathcal{G}$ with a node (vertex) set $\mathcal{V}_\mathcal{G}$, which is connected by a corresponding edge set $\mathcal{E}_\mathcal{G}$. These edges define the forward pass in the graph. The representation of a node differs based on the structure and design of the search space. Fine-grained methods like AIO-P~\cite{mills2023aiop} and AutoGO~\cite{salameh2024autogo} use an Intermediate Representation (IR) extracted from frameworks like ONNX~\cite{bai2019Onnx}, where nodes represent single primitive operations like convolutions, adds and activations. On the other hand, cell-based NAS-Benchmarks~\cite{mehta2022suite} use nodes or edges to represent predefined operation sequences. The coarsest node representation resides in macro-search spaces. Nodes could represent a whole residual convolution layer~\cite{he2016deep}, a transformer block with multi-head attention~\cite{vaswani2017attention, Rombach_2022_CVPR}, or MBConv~\cite{howard2017mobilenets, sandler2018mobilenetv2, howard2019searching, wan2020fbnetv2, Dai2021FBNetV3JA, tan2019efficientnet} block structure. Nodes are stacked in a 1-dimensional sequence to form a network.

We primarily focus on macro-search sequence graphs in this paper. Generally, a macro-search space contains two tiers of granularity: Stages (also known as Units~\cite{cai2020once, mills2021profiling}) and Layers. Architectures in a search space contain a fixed number of stages, each assigned a specific latent tensor Height-Width-Channel (HWC) dimension. Stages could contain a variable number of layers. Each layer constitutes a node that defines the specific type of computation such as convolution vs. attention in Stable Diffusion v1.4 or kernel size/channel expansion ratio if the layer is an MBConv block.

Each stage is denoted as a subgraph of its layers. The maximum number of unique subgraphs for a given stage depends on the number of layers it can contain as well as the set of layer type combinations that are permitted. Therefore, the size of the overall search space depends on the number of stages as well as the number of possible subgraphs for each stage. We provide an illustration of a sequence-like graph for a macro-search space in  Figure~\ref{fig:hops}, showing how nodes can be labeled with stage/layer positional features.

\subsection{Induced Subgraph Embeddings}
Graph Neural Networks (GNN)~\cite{you2020design} facilitate message passing between nodes using a graph's adjacency matrix. Each GNN layer represents a `hop' - where each node receives information from all immediate nodes in its neighborhood. Given node $v$, let $\mathcal{N}(v) = \{s \in \mathcal{V}_{\mathcal{G}} | (s, v) \in \mathcal{E}_\mathcal{G}\}$ denote the neighborhood of $v$. Furthermore, let $h_v^m \in \mathbb{R}^{1 \times d}$, where $d$ is the embedding vector length, represent the latent representation of $v$ at layer $m$. Then, the message passing function~\cite{xu2019GIN} of a GNN layer $m$ can be defined as: 

\begin{figure}[t]
    \centering
    \includegraphics[width=3.25in]{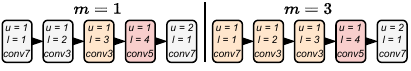}
    \caption{Visualization of a sequence-like architecture DAG. Nodes annotated with position and layer information: `u' and `l' refer to stage and layer position, while `conv' refers to the type of layer. Left graph: Subgraph rooted at the red node induced by a 1-hop message passing layer. Right graph: Additional (orange) nodes incorporated into the subgraph for a 3-hop layer.} 
    \label{fig:hops}
    \vspace{-4mm}
\end{figure}

\begin{equation}
   \centering
   \label{eq:gnn}
   h_v^m = \texttt{Combine}(\texttt{Agg}(h_v^{m-1}, \{h_s^{m-1}: s \in \mathcal{N}(v)\})
\end{equation}
where the \texttt{Combine} and \texttt{Agg} (Aggregate) functions are determined by GNN type, e.g., GAT~\cite{velivckovic2017graph, brody2022how}, GCN~\cite{welling2016semi}, etc.

With each successive GNN layer, the receptive field of $v$ grows as features from more distant nodes are aggregated into $h_v^m$. As such, instead of interpreting $h_v^m$ as the `embedding of node $v$ at layer $m$', we extend $h_v^m$ to denote the `embedding of the $m$-hop rooted-subgraph induced by node $v$'. Figure~\ref{fig:hops} illustrates how the subgraph encompasses information from all nodes within an $m$-hop distance of $v$.

\vspace{0.5em}
\noindent \textbf{Graph Aggregation and Prediction.}
In graph-level prediction problems, the GNN produces a single prediction per graph. Since graphs can have varying number of nodes and edges, this requires condensing the node embeddings from a message passing layer 
$H_{\mathcal{V}_{\mathcal{G}}}^m 
\in \mathbb{R}^{|\mathcal{V}_{\mathcal{G}}| \times d}$ 
into a single, fixed-length embedding representing the \textit{whole} graph, $h_{\mathcal{G}}^m \in \mathbb{R}^{1 \times d}$. Typically, this is performed by applying an arithmetic operation like `mean'~\cite{you2020design}, 

\begin{equation}
    \centering
    \label{eq:aggr}
    h_{\mathcal{G}}^m = \dfrac{1}{|\mathcal{V}_{\mathcal{G}}|}\sum_{v \in \mathcal{V}_{\mathcal{G}}} h_v^m. 
\end{equation}
A GNN can contain multiple layers $m \in [0, M]$. Usually, a single graph embedding is calculated from the output of the final message passing layer $M$, which is then fed into a simple Multi-Layer Perceptron (MLP) to produce a prediction, $y'_{\mathcal{G}} = \texttt{MLP}(h_{\mathcal{G}}^M)$. However, it is possible to calculate subgraph embeddings at \textit{any} hop-level, e.g., $m < M$, including $m=0$. In fact, it is this capability which allows AutoBuild to `score' and compare subgraphs of different sizes as well as node feature categories, as we will next discuss. 

\section{Methodology}
\label{sec:method}

In this section, we elaborate on the design of AutoBuild, specifically the hop-level correlation loss, and how we can use a Magnitude Ranked Embedding Space to compare subgraphs induced by different hop levels. Additionally, we elaborate on our Feature Embedding MLP (FE-MLP). 

\subsection{Magnitude Ranked Embedding Space}
Graph regressors learn to estimate their target values by minimizing loss, such as the Mean-Squared Error (MSE) or Mean Absolute Error (MAE) between its predictions $y'_{\mathcal{G}}$ and targets $y_{\mathcal{G}}$. However, this approach does not impose any constraint upon the latent embedding space which makes the predictor a black box with hard to interpret  predictions. On the other hand, AutoBuild constrains the embedding space by associating high-performance subgraph stages with high scores, making it easier to interpret results.

Let $\mathcal{G}_1$ and $\mathcal{G}_2$ be two separate architecture graphs in a training set. In addition to a standard regression loss between labels and targets, AutoBuild incorporates an additional constraint into the learning process:

\begin{equation}
    \centering
    \label{eq:correlation}
    \resizebox{0.433\textwidth}{!}{
    $\forall m \in [0, M] \ |\ \texttt{if}\ y_{\mathcal{G}_1} > y_{\mathcal{G}_2}, \texttt{then}\ \norm{h^{m}_{\mathcal{G}_1}}_1 > \norm{h^{m}_{\mathcal{G}_2}}_1.$}
\end{equation}
This constraint ensures that if the ground-truth label for $\mathcal{G}_1$ is higher than $\mathcal{G}_2$, then the norm of $\mathcal{G}_1$'s graph embedding should be higher than that of $\mathcal{G}_2$ across all hop levels. In practice, we impose this constraint by augmenting the original regression loss with an additional ranking loss,

\begin{equation}
    \centering
    \label{eq:loss}
    \mathcal{L} = \texttt{MSE}(y_{\mathcal{G}}, y'_\mathcal{G}) + \dfrac{1}{M+1}\sum_{m = 0}^M(1 - \rho(y_\mathcal{G}, \norm{h^{m}_{\mathcal{G}}}_1)),
\end{equation}
where $\rho$ is the Spearman's Rank Correlation Coefficient (SRCC). The value of $\rho = 1$ indicates perfect positive correlation, while $\rho = -1$ indicates negative correlation and $\rho = 0$ denotes no correlation. If the graph embedding norms are perfectly aligned with the targets, $\rho(y_\mathcal{G}, \norm{h^{m}_{\mathcal{G}}}_1) = 1$, the loss term will be zeroed. Typically, ranking metrics like SRCC are generally considered non-differentiable and are therefore not usable as loss functions. However, we utilize the method of Blondel et al., 2020~\cite{blondel2020fast}, which provides an exact, differentiable computation to calculate SRCC across an entire batch of data.

Equation~\ref{eq:loss} forces the GNN to learn embeddings in a Magnitude Ranked Embedding Space - where graph embedding norms are correlated with ground-truth targets. Moreover, since graph embeddings $h_\mathcal{G}^m$ are directly computed from node embeddings $h_v^m$ per Equation~\ref{eq:aggr}, the GNN is forced to learn which nodes can be attributed to these more prominent labels and subsequently assign them larger node embeddings. Thus, the norm of the node embedding, $\norm{h_v^m}_1$, acts as a learned numerical `score' for the node $v$, and by extension, its induced $m$-hop subgraph.

We further note that the node/subgraph scores do not require additional architecture annotations, e.g., explicit ground-truth subgraph labels. Rather subgraph scores are implicitly learned from the training data. Like other neural performance predictors~\cite{white2021powerful}, the only ground-truth architecture annotations AutoBuild requires to compute Eq.~\ref{eq:loss} is $y_\mathcal{G}$, the scalar label for architecture $\mathcal{G}$. $y_\mathcal{G}$ is typically an end-to-end architecture metric like accuracy, inference latency, or a combination thereof, as we demonstrate in Sec.~\ref{sec:results}.

\subsection{Comparing Subgraphs at Different Hops}
We can interpret the embedding norm $\norm{h^m_v}_1$ as the score for the $m$-hop subgraph rooted at node $v$. However, this score is only comparable between subgraphs of the same hop level. 
While Equation~\ref{eq:loss} ensures ranking correlation between embedding norms and the ground-truth labels, the distribution of embedding norms can vary \textit{between} hop levels. Therefore, additional computation is necessary to compare subgraphs of different sizes. We propose a mechanism that works in the case of macro-search spaces where architectures can consist of stages (Sec~\ref{sec:arch_formats}), since each stage can consist of one predefined subgraph. 

Let $\mu^h_{m}$ and $\sigma^h_{m}$ refer to the mean and standard deviation of node embedding norms produced by the GNN at hop-level $m$. Also, let $\mu^y_{u, l}$ and $\sigma^y_{u, l}$ refer to the mean and standard deviation of ground-truth labels for architectures that have $l$ layers ($l = m+1$). To compare an arbitrary node embedding norm $\norm{h^m_v}_1$ with a score from another hop-level, we shift it from $\mathcal{N}(\mu^h_{m}, \sigma^h_{m})$ to $\mathcal{N}(\mu^y_{u, l}, \sigma^y_{u, l})$ as follows: 

\begin{equation}
    \centering
    \label{eq:distshift}
    \norm{h^l_v}_1^* = (\norm{h^m_v}_1 - \mu^h_m) * \dfrac{\sigma^y_{u, l}}{\sigma^h_m} + \mu^y_{u, l},  
\end{equation}
where $\norm{h^l_v}_1^*$ is the shifted embedding norm. Likewise, in order to compare this subgraph to another one at a different hop-level, e.g., $\norm{h^{m+1}_v}_1$, we would in turn shift that score using $\mathcal{N}(\mu^h_{m+1}, \sigma^h_{m+1})$ and $\mathcal{N}(\mu^y_{u, l+1}, \sigma^y_{u, l+1})$ instead.

This distribution shift aligns the subgraph scores with the biases of the target data. While we could standardize all scores into a normal distribution $\mathcal{N}(0, 1)$, such a procedure assumes that subgraphs at different hop levels should be weighed equally. For example, standardizing prior to comparison assumes that the average $l$-hop subgraph should receive the same score as the average $l+1$-hop subgraph. However, if the GNN is an accuracy or latency predictor, using a bigger subgraph at a given stage means the overall architecture is larger, which on average entails higher accuracy and latency ~\cite{mills2021profiling}. As such, we would expect $\mu^y_{u, l+1}$ to exceed $\mu^y_{u, l}$. We can interpret the difference $\mu^y_{u, l+1} - \mu^y_{u, l}$ as the performance \textit{bias} an $l+1$-sized stage subgraph enjoys, and that a smaller $l$-hop stage subgraph must overcome, through superior combination of layers, to obtain a higher score. Therefore, we first standardize using $\mathcal{N}(\mu^h_{m}, \sigma^h_{m})$, which is computed by passing the graph dataset over the GNN after training is complete. We then unstandardize using $\mathcal{N}(\mu^y_{u, l}, \sigma^y_{u, l})$, which can be statistically computed from the training set.

\subsection{Ranking Individual Node Features}

The first component of a GNN is typically the embedding layer, $m=0$, which translates the original, human-readable features of graph nodes into numerical vectors that machine-learning models can better understand. The nodes can have various types of features, such as discrete operation types like `convolution', or numerical vectors for tensor shapes. Positional encodings like the stage $u$ and layer $l$ are also considered. The GNN embedding layer is responsible for refining these different feature categories into a single continuous vector, $h^0_v \in \mathbb{R}_{+}^{1 \times d}$.

AutoBuild learns to estimate the importance of individual node feature categories by incorporating a Feature Embedding Multi-Layer Perception (FE-MLP). The design of the FE-MLP is straightforward: each node feature category is assigned a separate feed-forward module that processes it into a continuous scalar. We then concatenate the scalar values for all feature categories, and apply an \texttt{abs} non-linearity to form $h^0_v$, the `0-hop' node embedding. 

Although $h^0_v$ consists of scalars from all feature categories, they do not interact with each other. By coupling this design choice with the hop-level ranking loss in Equation~\ref{eq:loss}, the FE-MLP can identify which feature choices should be assigned high importance scores without any additional information about the given node. This allows us to easily determine the importance of each node feature choice once training is complete, without having to exhaustively evaluate every possible node feature combination.

\section{Results}
\label{sec:results}

We gauge the effectiveness of AutoBuild in several domains. First, we find Pareto-optimal architectures on two ImageNet-based macro-search spaces from OFA. Then, we apply AutoBuild to two domains where we have access to a limited amount of labeled architectures: Panoptic Segmentation for MS-COCO~\cite{coco} and Inpainting using SDv1.4. For all experiments, we provide a full description of the training setup and hyperparameters in the supplementary materials.

\begin{figure}[t]
    \centering
    \subfloat{\includegraphics[width=3.2in]{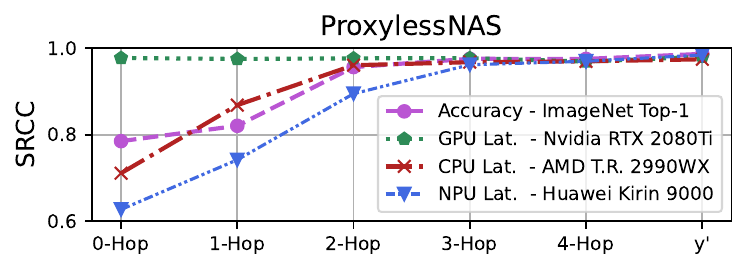}}
    \qquad
    \subfloat{\includegraphics[width=3.2in]{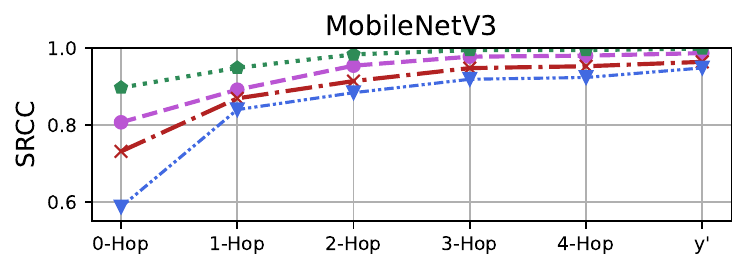}}
    \caption{Test SRCC for PN and MBv3 test sets. Specifically, we train AutoBuild predictors using Eq.~\ref{eq:loss} for accuracy or inference latency on several hardware devices, then measure SRCC for every hop level $m \in [0, 4]$ and the predictor MLP (y').} 
    \label{fig:ofa_srcc}
    \vspace{-4mm}
\end{figure}

\subsection{ImageNet Macro-Search Spaces}
\label{sec:srcc}

Our OFA macro-search spaces are ProxylessNAS (PN) and MobileNetV3 (MBv3). Specifically, each search space follows the stage-layer layout described in Sec~\ref{sec:arch_formats}. Both PN and MBv3 contain over $10^{19}$ architectures and use MBConv blocks as layers. There are 9 types with kernel sizes $k \in \{3, 5, 7\}$ and channel expansion ratio $e \in \{3, 4, 6\}$. We further enrich the node features by two position-based node features that encode a node stage $u$ and layer $l$ location. MBv3 contains 5 stages each with 2-4 layers. Thus, the total number of architecture module subgraphs for MBv3 is $9^2+9^3+9^4 = 7.4k$. PN has $9$ more modules as it includes a final 6th stage that always has 1 layer. 

To obtain accuracy labels, we sample and evaluate $3k$ architectures from each search space. Additionally, we obtain inference latency measurements from \cite{mills2021profiling} for three devices: RTX 2080 Ti GPU, AMD Threadripper 2990WX CPU, and Huawei Kirin 9000 Mobile NPU~\cite{Liao2021AscendAS}.

We gauge the effectiveness of Equation~\ref{eq:loss} by training GNN predictors and then measuring the test set SRCC of the graph embeddings at each hop-level. For the training/testing data split, we use a ratio of 90\%/10\% and a batch size of 128. Figure~\ref{fig:ofa_srcc} plots the SRCC values across each search space and target metric. We observe strong positive SRCC values across all hop levels for every search space and target metric. This demonstrates the feasibility of learning a magnitude-ranked embedding space using Eq.~\ref{eq:loss}.

\begin{figure}[t]
    \centering
    \subfloat{\includegraphics[width=1.63in]{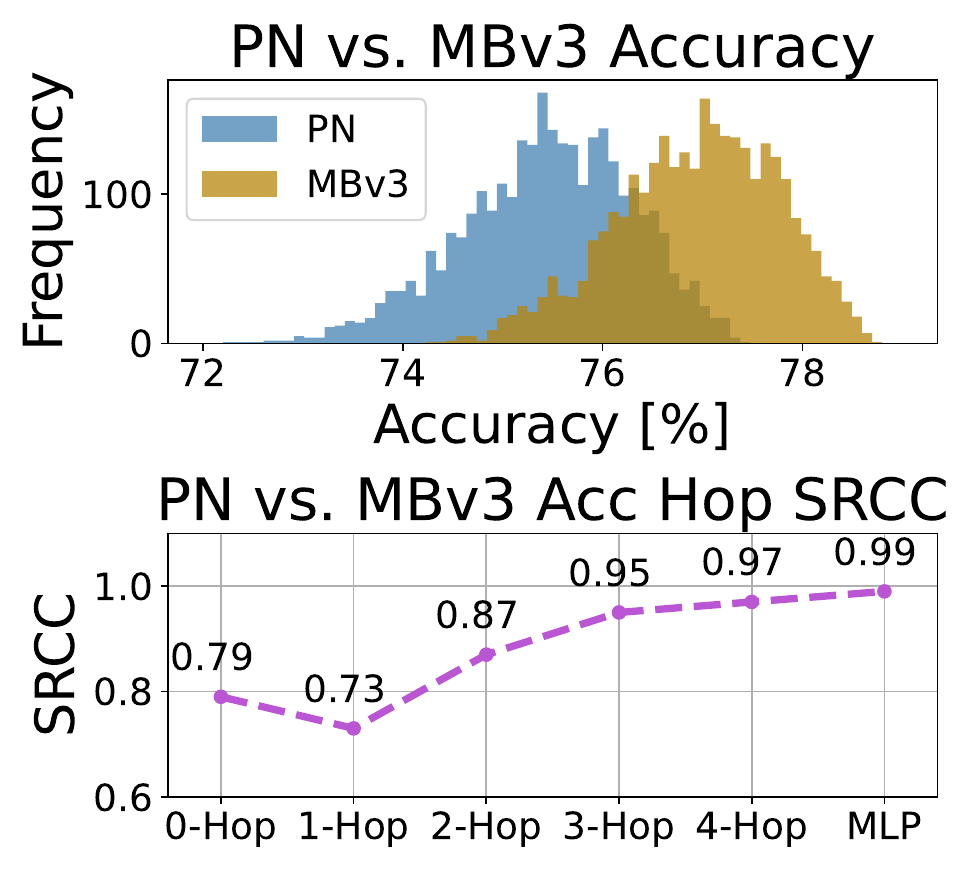}}  
    \subfloat{\includegraphics[width=1.65in]{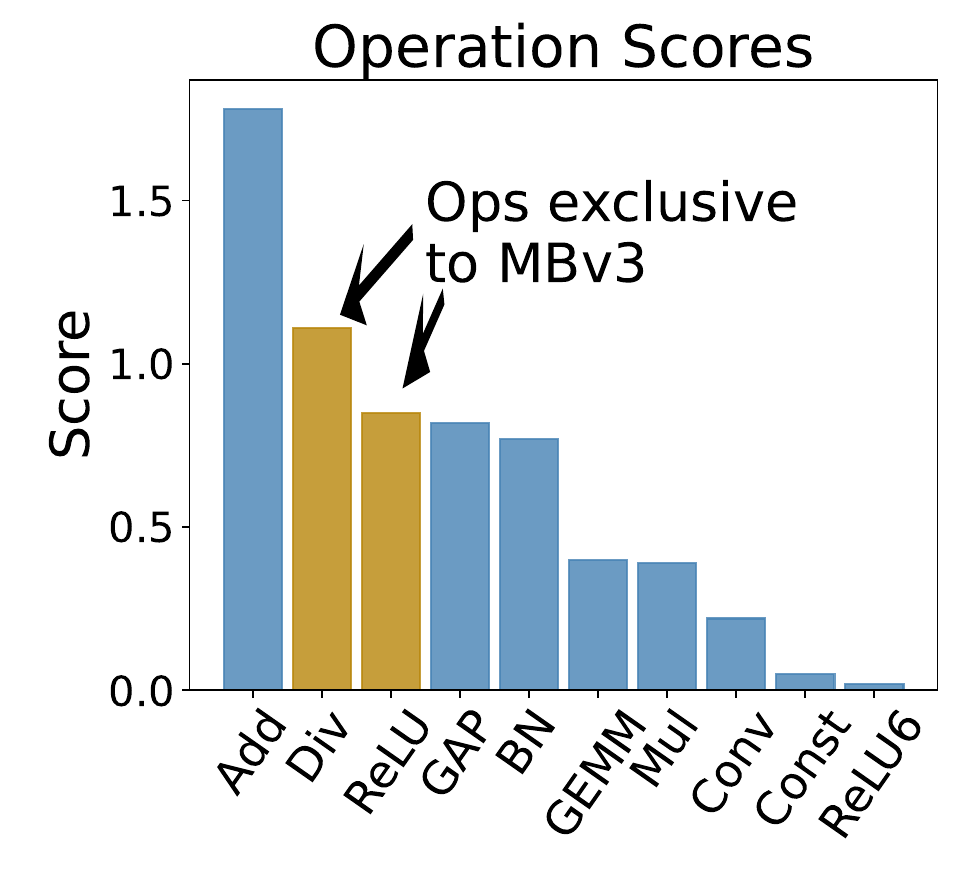}} 
    \caption{Results of training an accuracy predictor for PN and MBv3 using the `IR' representation format. \textbf{Top-left}: Accuracy histogram. \textbf{Bottom-left}: Hop-wise AutoBuild test SRCC. \textbf{Right}: Operation-type importance scores from the 0-hop FE-MLP. Note: `GAP' means Global Average Pool and `BN' means Batch Norm.}
    \label{fig:mobilenet_onnx_ir}
    \vspace{-4mm}
\end{figure}

\vspace{0.5em}
\noindent\textbf{Cross-Search Space Comparison.}
To illustrate how the FE-MLP can assign human-interpretable importance scores to node-level features, we train an accuracy predictor for PN and MBv3 using the finer ONNX IR graph format mentioned in Sec.~\ref{sec:arch_formats}. Unlike sequence graphs, the IR format represents operation primitives such as conv or ReLU as nodes while 
operation type is a categorical node feature.

We present our findings in Figure~\ref{fig:mobilenet_onnx_ir}. The 0-hop SRCC of the predictor is $0.79$ so FE-MLP embedding norms are correlated with accuracy. First, note MBv3 has a higher accuracy distribution than PN. The FE-MLP captures this distinction, by assigning high importance to `Division' and `ReLU' operation nodes, which are part of the \textit{h-swish} activation~\cite{howard2019searching} and Squeeze-and-Excite module (SE)~\cite{hu2018squeeze} exclusive to MBv3 since PN only uses the `ReLU6' activation. Additionally, the operation with the highest FE-MLP score is the `Add', due to its multi-input nature and the frequent use of long skip-connects. We provide a detailed analysis of this phenomenon in the supplementary material.

\begin{figure}[t]
    \centering
    \subfloat[$y=100^{Acc}$]{\includegraphics[width=3.25in]{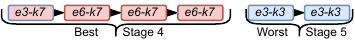}}
    \vspace{1mm}
    \qquad
    \subfloat[$y=100^{Acc}/log_{10}(Lat)$]{\includegraphics[width=3.25in]{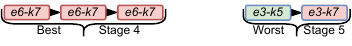}}
    \vspace{1mm}
    \qquad
    \subfloat[$Acc/Lat$]{\includegraphics[width=3.25in]{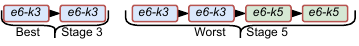}}
    \vspace{1mm}
    \qquad
    \subfloat[$\sqrt{Acc}/Lat$]{\includegraphics[width=3.25in]{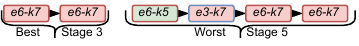}}
    \caption{Best and worst MBConv layer subgraphs for MobileNetV3 on GPU latency, annotated with the stage where they are found. Specifically, we illustrate the best and worst subgraphs, by AutoBuild module score, when different equations are used to compute the predictor targets using accuracy and latency. }
    \label{fig:subgraphs}
    \vspace{-4mm}
\end{figure}

\begin{figure*}[t]
    \centering
    \subfloat{\includegraphics[height=0.84in]{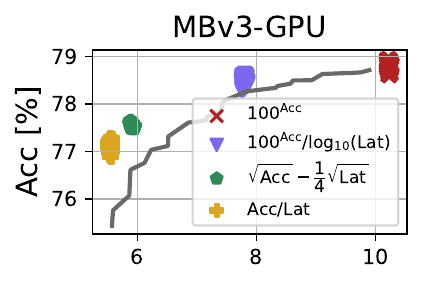}}
    \subfloat{\includegraphics[height=0.84in]{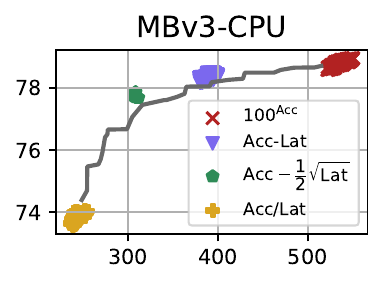}}
    \subfloat{\includegraphics[height=0.84in]{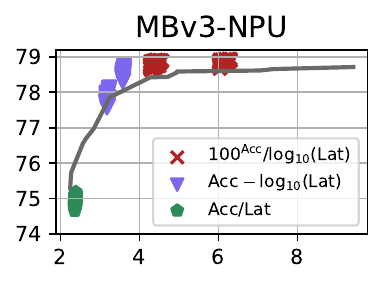}}
    \subfloat{\includegraphics[height=0.84in]{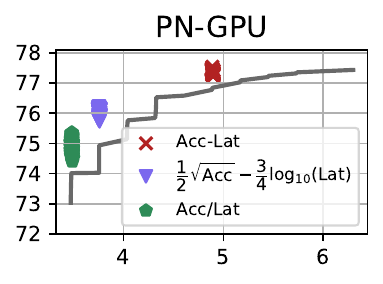}}
    \subfloat{\includegraphics[height=0.84in]{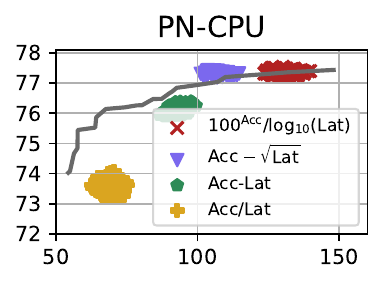}}
    \subfloat{\includegraphics[height=0.84in]{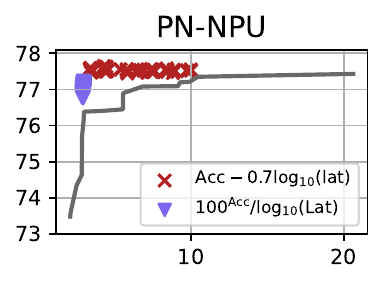}}
    \qquad
    \subfloat{\includegraphics[height=0.84in]{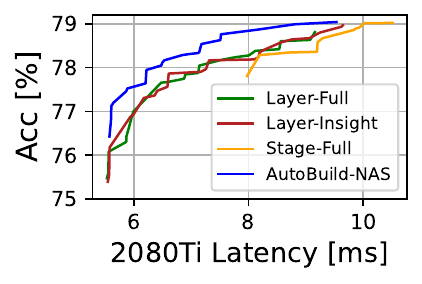}}
    \subfloat{\includegraphics[height=0.84in]{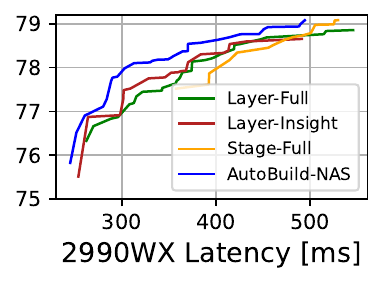}}
    \subfloat{\includegraphics[height=0.84in]{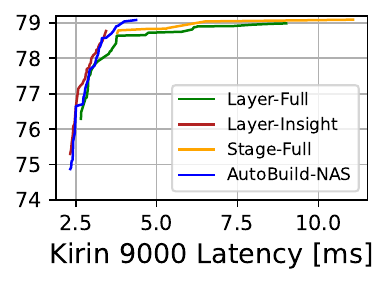}}
    \subfloat{\includegraphics[height=0.84in]{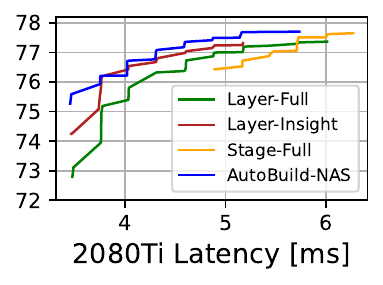}}
    \subfloat{\includegraphics[height=0.84in]{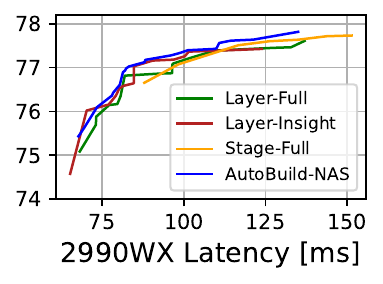}}
    \subfloat{\includegraphics[height=0.84in]{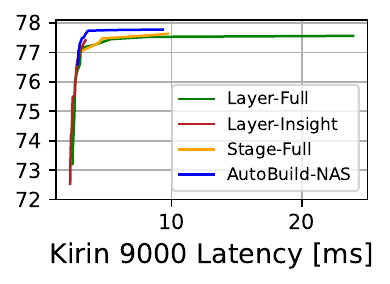}}
    \caption{Top row: Comparing reduced search spaces produced by AutoBuild (colored clusters; K=3) to the accuracy-latency Pareto frontier of the architectures used to train the predictor (grey line). Each cluster corresponds to a specific target equation designed to improve upon a specific region of the frontier. Bottom row: Evolutionary search results comparing the Unit-Reduce search space (K=25) of AutoBuild to other search spaces and mutation techniques. Best viewed in color.}
    \label{fig:paretos}
    \vspace{-4mm}
\end{figure*}

\vspace{0.5em}
\noindent\textbf{Architecture Construction.}
Next, we demonstrate how module subgraph-scoring can help us construct a small handful of high-efficiency architectures. Specifically, we build architectures that outperform the accuracy-latency Pareto frontier formed by the $3k$ random architectures we use to train and test our predictors.

First, we plot the accuracy vs. latency Pareto frontiers for existing architectures. Then, using these architectures, we train several AutoBuild predictors using different target label equations that combine both accuracy and latency. These equations aim to optimize different regions of the dataset Pareto frontier by emphasizing accuracy and latency to different degrees, e.g., $y=\sfrac{acc}{lat}$, $y=acc - lat$, etc. 

We then compute the AutoBuild score for each architecture module subgraph in the search space. Figure~\ref{fig:subgraphs} illustrates the best and worst subgraphs for MBv3 on GPU latency for different label equations. These results are intuitive, e.g., for $y=100^{Acc}$, the best subgraph has 4 layers (the maximum),  convolving over many channels ($e6$) using the largest possible kernel size ($k7$), while the worst subgraph is small with few channels and low kernel size. When latency is factored in, the best subgraphs contain fewer layers yet maintain high expansion $e$ values, while bad subgraphs are long sequences containing inefficient blocks. These results align with \cite{mills2021profiling} who found that MBv3 GPU latency was primarily driven by the number of layers per stage rather than layer type. 

Next, using the AutoBuild scores, we iterate across each stage $u$ and select the top-$K$ subgraphs with the highest scores, forming a reduced search space with size $K^{|u|}$. Individual architectures that optimize the predictor target can be quickly constructed by sampling one module subgraph per stage. If $K$ is small, e.g., $K=3$ while $|u|=5$, then $K^{|u|}=243$. Using OFA~\cite{cai2020once} supernetworks, a search space of this size can be exhaustively evaluated in a few hours on a single GPU. Afterwards, we plot the accuracy and latency of the constructed architectures against the existing architecture Pareto frontier.

The top row of Figure~\ref{fig:paretos} illustrates our findings for MBv3 and PN. On MBv3, PN-GPU and PN-NPU we consistently craft target equations that can focus on regions that contain architectures beyond the Pareto frontier. On MBv3-CPU, MBv3-NPU and PN-NPU, we achieve better performance then architectures on the high-accuracy tail of the frontier using significantly less latency. On PN-CPU, only two of the four reduced search spaces outperform the data Pareto frontier. We note that our equations are manually inferred from the data which is a limitation. Automating the equation-creation process is a future research direction.

\vspace{0.5em}
\noindent\textbf{Enhancing Neural Architecture Search.} 
AutoBuild is not limited to simply constructing a handful of specialized architectures. By combining the top-$K$ architecture module subgraphs from predictors trained using different target equations, we can craft a reduced search space. Further, we can then use a NAS algorithm to find a Pareto frontier of high-efficiency architectures.

Specifically, for target equations in the top row of Fig.~\ref{fig:paretos}, we set $K=25$, then combine the set of module subgraphs selected for each stage to form our search space. We perform NAS using a multi-objective evolutionary search algorithm~\cite{liu2021EvoSurvey, sun2020Evo, lu2019nsga} which performs mutation by randomly swapping unit subgraphs. We denote our approach as \textbf{AutoBuild-NAS} and compare it to several baselines: 

\begin{itemize}
    \item \textbf{Stage-Full} uses the whole search space. This baseline performs mutation by changing one of the stage subgraphs in an architecture.
    \item \textbf{Layer-Full}: Similar to Stage-Full with a different mutation mechanism that adds, removes or edits a single layer at a time, instead of changing whole stage subgraphs. 
    \item \textbf{Layer-Insight}: Layer-Full mutation using the reduced search spaces of \cite{mills2021profiling}, who generate insights by statistically sampling the impact of each \textit{stage-layer} combination using many architectures. Manual, human-expert knowledge is applied to restrict layer choice and the maximum number of layers per stage. 
\end{itemize}

We set a search budget of 250 architectures. In the supplementary materials, we provide additional search details and calculate the size of each method's search space. The bottom row of Figure~\ref{fig:paretos} illustrates our search results. We note that AutoBuild-NAS leads to superior Pareto frontiers compared to other methods across all search spaces and hardware devices, specifically all MBv3 tests and PN-GPU. In fact, while both AutoBuild-NAS and Stage-Full can break 79\% accuracy on MBv3-CPU and MBv3-NPU, only AutoBuild-NAS achieves this at less than 5ms latency on MBv3-NPU. Also, at the lower-latency end of the Pareto frontiers, the strongest competitor is generally Layer-Insight, who also shrink the search space. However, gains attained by Layer-Insight are quite small and inconsistent. Moreover, while Layer-Insight sampled many thousands of architectures to build insights and relies on upon human knowledge to decide how to reduce the search space, AutoBuild-NAS is data-driven and only relies on $3k$ labeled samples to train each AutoBuild predictor. 

\subsection{Application to Panoptic Segmentation}
\label{sec:pan_seg_aiop}

\begin{figure}
    \centering
    \includegraphics[height=1in]{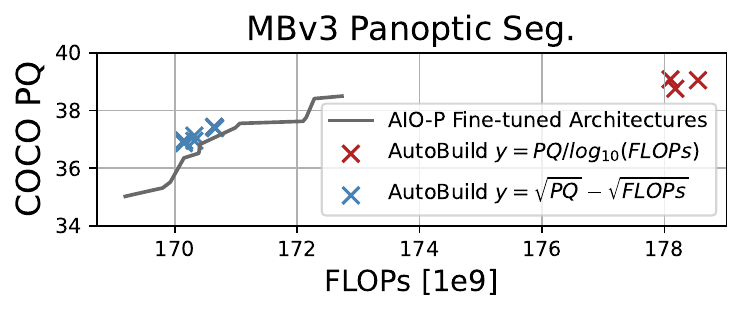}
    \caption{Results comparing AutoBuild MBv3 architectures to the PQ-FLOPs Pareto frontier of fine-tuned architectures from \cite{mills2023aiop}.} 
    \label{fig:pck_pq}
    \vspace{-4mm}
\end{figure}

AutoBuild can construct architectures for higher-resolution computer vision tasks, like Panoptic Segmentation (PS)~\cite{wu2019detectron2, kirillov2019panoptic} on MS-COCO~\cite{coco}. However, evaluation on this task is costly compared to classification as it requires fine-tuning on over 50GB of VRAM for over 24 V100 GPU hours per MBv3 architecture. AIO-P~\cite{mills2023aiop} provides performance annotations for ${\sim}100$ fully fine-tuned and ${\sim}1.3k$ `pseudo-labeled' MBv3 architectures, respectively. 

We train AutoBuild predictors to find MBv3 module subgraphs that provide superior and efficient Panoptic Quality (PQ). Specifically, we calculate module scores by training predictors on the `pseudo-labeled' architectures using two target equations: $PQ/log_{10}(FLOPs)$ aims to maximize PQ while not finding overtly large architectures, while $\sqrt{PQ}-\sqrt{FLOPs}$ strikes a balance between performance and computation cost. For each equation, we evaluate a small handful of constructed architectures, then compare to their PQ and FLOPs to the Pareto frontier derived from the fully fine-tuned architectures.

Figure~\ref{fig:pck_pq} presents our findings. First, we note that module subgraphs selected using $y=PQ/log_{10}(FLOPs)$ obtain high PQ, generally at or above 39 which is better than any of the fully fine-tuned architectures from \cite{mills2023aiop}. However, the FLOPs of these architectures is quite high and generally overshoots the high-FLOPs tail of the Pareto frontier by about $5$ billion FLOPs. By contrast, architectures constructed using $y=\sqrt{PQ}-\sqrt{FLOPs}$ strike a balance between performance and the computational burden and outperform the existing Pareto frontier. Thus, Fig.~\ref{fig:pck_pq} demonstrates the applicability of AutoBuild in scenarios where evaluation is costly and the ability to use `pseudo-labeled' samples to provide accurate architecture module scores.

\subsection{Generative AI with Limited Evaluations}
\label{sec:inpainting}

AutoBuild is applicable in emerging Generative AI tasks, such as Inpainting based on Stable Diffusion, under very limited architecture evaluations. In this task, we mask out certain parts of an image and rely on Stable-Diffusion models to refill the missing part. Training diffusion models involve a forward pass where Gaussian noises are added to the input image in a step-wise fashion. The masked area is replaced during inference with Gaussian noise, so only the reverse process is required to generate new content in that region. The baseline Inpainting model, SDv1.4~\cite{Rombach_2022_CVPR}, consists of a VQ-GAN~\cite{esser2021taming} encoder/decoder and a denoising U-Net which has 4 stages, each consisting of a series of residual convolution and attention layers. We illustrate the whole model structure in the supplementary.

Our goal is to find a high-performance variant of the original SDv1.4 U-Net by altering the number of channels in each stage and enabling/disabling attention or convolution layers in each stage. Similar to SnapFusion~\cite{li2023snapfusion}, we remove attention from the early stages of the network, as they require too many FLOPs and do not provide much performance gain. The above pruning and channel altering forms a macro-search space containing ${\sim}800k$ architectures.

To apply AutoBuild, we first generate a set of $68$ random architectures from this search space including SDv1.4 U-Net, each inheriting weights from the original SDv1.4 U-Net. We fine-tune architectures on the Places365~\cite{zhou2017places} training set for 20k steps, then measure the Fr\'echet Inception Distance (FID)~\cite{heusel2017gans} using 3k images from the validation set with predefined masks. Each architecture fine-tuning (and the evaluation) takes 1-2 days and over 60GB of VRAM. As such, traditional NAS methods would require several dozens of GPUs to explore $0.1\%$ of 800k architectures. We show that just based on these 68 evaluated architectures, AutoBuild can directly build better architectures.

To deal with a small number of evaluated architectures, we train an ensemble of $20$ AutoBuild predictors, each trained on 4/5 of the 68 architectures under a 5-fold split. This procedure is repeated under 4 random seeds (to shuffle data), leading to 20 predictors. Then, each subgraph, which is a specific sequence of convolution and attention layers with a stage label and other features including number of channels, is scored by a weighted sum of this subgraph's embedding norms under the 20 predictors, where the weighting is given by the test SRCC between the subgraph embedding norms by a predictor and ground-truth labels at the hop-level of the subgraph in question. Thus, we can use the test SRCC on the remaining 1/5 held-out architectures as a confidence measure to weigh a predictor's estimation of subgraph importance. For AutoBuild, we fine-tune and evaluate the top-4 architectures according to the summation of weighted subgraph scores over each stage.

Additionally, we also consider a baseline search strategy, which applies Exhaustive Search (ES) to all 800k architectures, with performance estimation provided by an ensemble of 20 predictors trained in the same way as above except that only the MSE loss (between predicted and ground-truth FID) is used in training predictors since no node/subgraph embeddings are generated. The other difference is that the predictor weight is the test SRCC between FID predictions and the ground-truth. We then fine-tune and evaluate the top-4 architectures selected by ES.

\begin{table}[t]
    \centering
    \caption{Comparing AutoBuild to an Exhaustive evaluation technique on SDv1.4 and the FID scores (lower is better) in the predictor dataset. Brackets $()$ denote architecture set size.} 
    \scalebox{0.77}{
    \begin{tabular}{lccc} \toprule
    \textbf{Arch Set} & \textbf{Eval Archs (68)} & \textbf{Exhaustive Search (4)} & \textbf{AutoBuild (4)} \\ \midrule
    Ave. FID & 22.13 & 10.82 & \textbf{10.13} \\
    Best FID & 10.54 & 10.29 & \textbf{9.96} \\ \bottomrule
    \end{tabular}
    }
    \label{tab:fid}
    \vspace{-5mm}
\end{table}

Table~\ref{tab:fid} shows the true FID scores for the top-$4$ architectures selected by each method. While both methods can find architectures that are better than the 68 evaluated architectures, AutoBuild finds much better architectures than ES does, on average, and in terms of the best architecture found (with a much lower FID). This demonstrates the power of AutoBuild in constructing good architectures through interpretable subgraph mining, which cannot be achieved by ensembled predictors trained on the same 68 evaluated architectures.
In addition, ES (with ensembled predictions) is only possible due to the search space being still relatively small ($800k$). Since feeding each candidate architecture into the predictor will still cause overhead, for even larger search spaces, ES becomes undesirable or infeasible. We provide additional details and results in the supplementary.

\begin{figure}[t]
    \centering
    \subfloat[Original]{\includegraphics[height=0.9in]{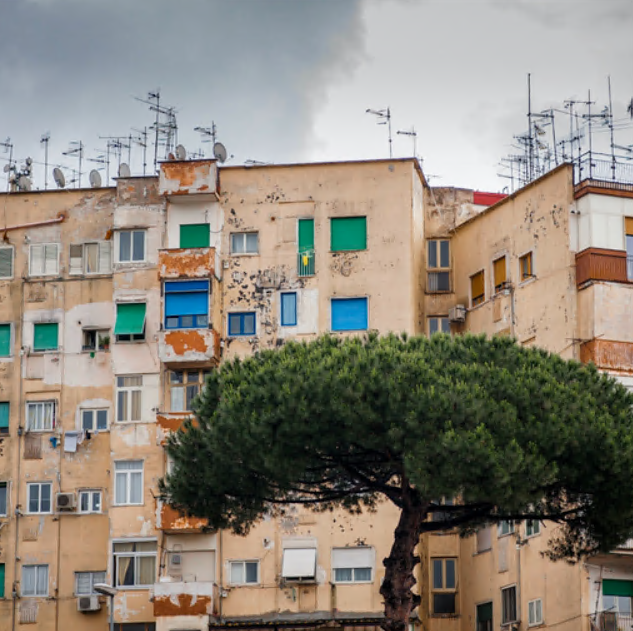}}
    \subfloat[ES]{\includegraphics[height=0.9in]{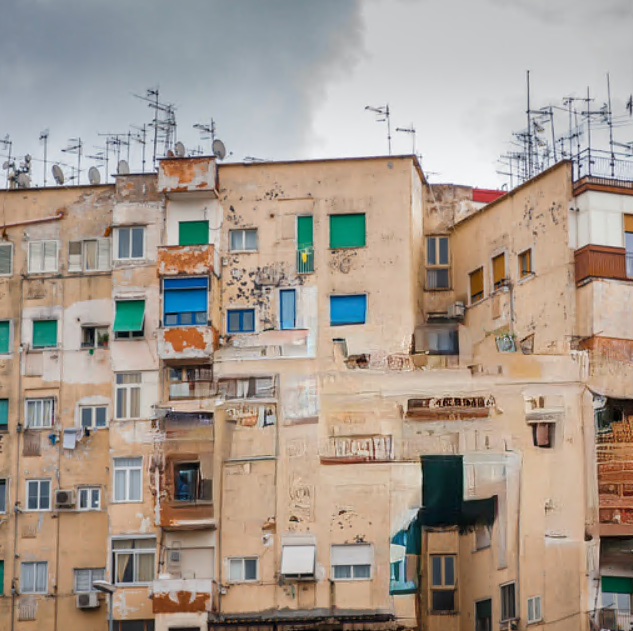}}
    \subfloat[AutoBuild]{\includegraphics[height=0.9in]{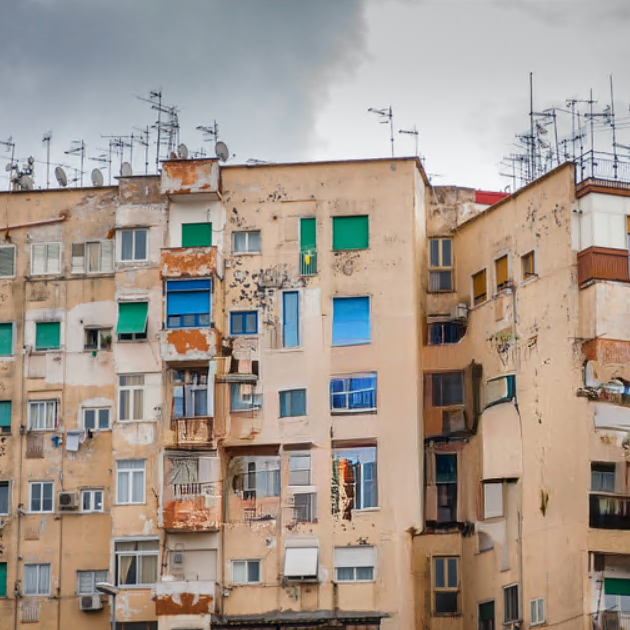}}
    \qquad
    \subfloat[Original]{\includegraphics[height=0.9in]{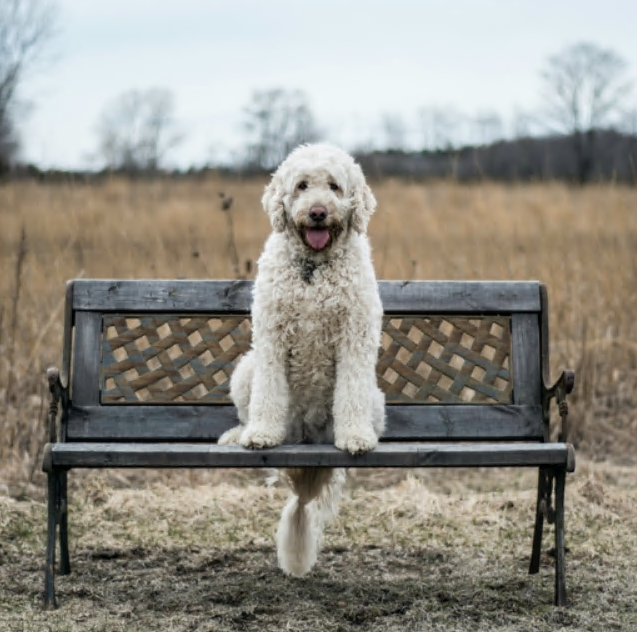}}
    \subfloat[ES]{\includegraphics[height=0.9in]{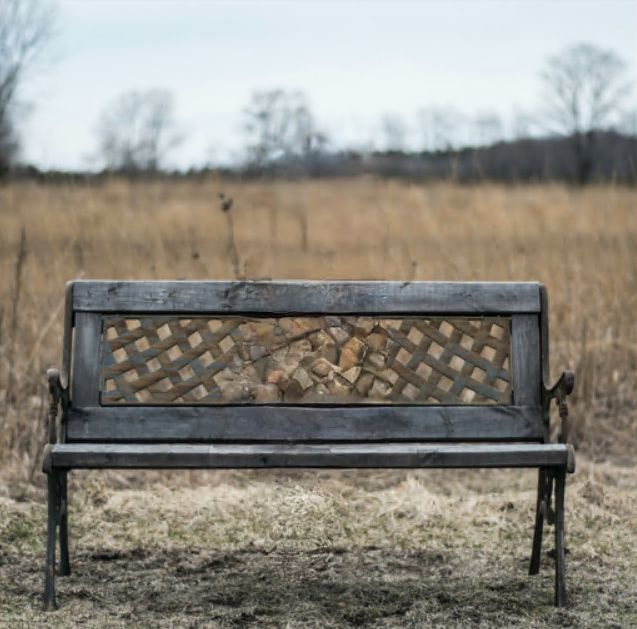}}
    \subfloat[AutoBuild]{\includegraphics[height=0.9in]{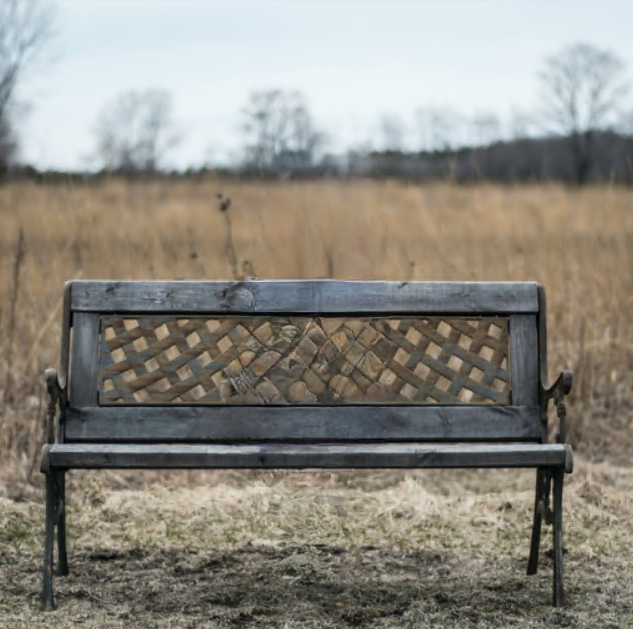}}
    \qquad
    \subfloat[Original]{\includegraphics[height=0.9in]{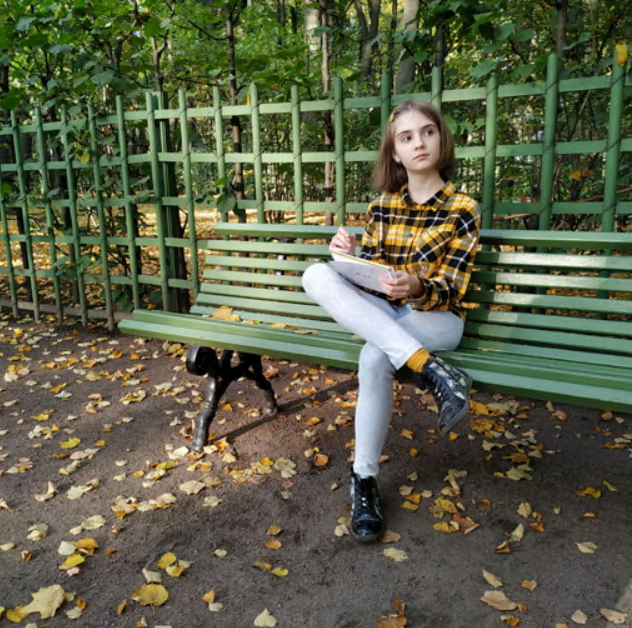}}
    \subfloat[ES]{\includegraphics[height=0.9in]{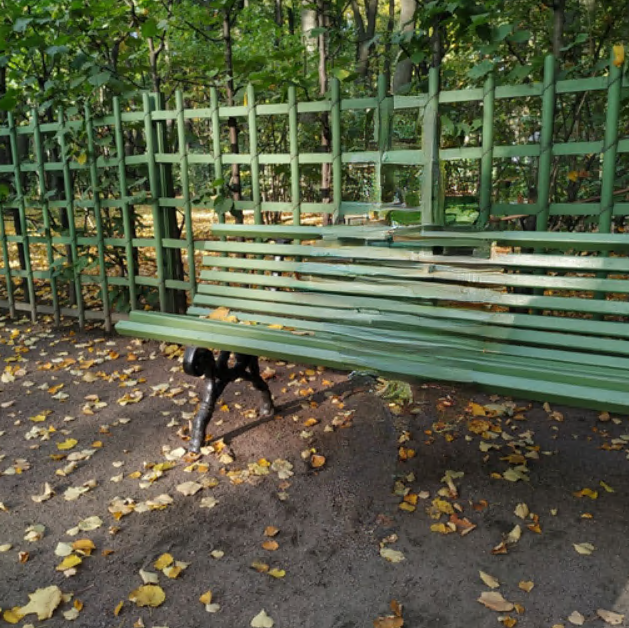}}
    \subfloat[AutoBuild]{\includegraphics[height=0.9in]{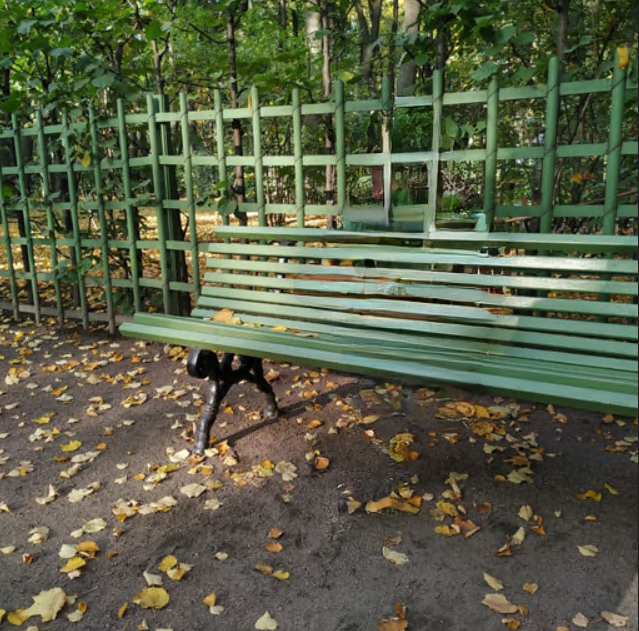}}
    \caption{Visual comparison of the best architectures found using exhaustive search and AutoBuild on sample images from SDv1.4.}
    \label{fig:inpainting_examples}
    \vspace{-4mm}
\end{figure}

Finally, we visually compare the inpainting outcomes of the best architectures found by AutoBuild and ES. Figure~\ref{fig:inpainting_examples} provides Inpainting samples from both U-Net variants. The AutoBuild architecture clearly produces more accurate content, e.g., faithfully representing the building structure and the backrests of both benches whereas the ES variant fails to do so. Thus, these findings demonstrate the robust utility of AutoBuild and architecture module scoring.

\section{Conclusion}
\label{sec:conclusion}

We propose AutoBuild, to measure and rank the importance of the architecture modules that exist within search spaces as opposed to focusing on traversing the whole design space as traditional NAS does. AutoBuild applies a hop-level ranking correlation loss to learn graph embedding norms that are correlated with architecture performance labels. This method allows us to assign numerical scores to architecture modules of different sizes. We then use these scores to quickly construct high-quality classification and  segmentation models to improve existing NAS procedures by reducing the search space. Finally, we show that AutoBuild is applicable in scenarios, where few architectures have been evaluated, as we use it to find a high-performance variant of Stable Diffusion v1.4 for Inpainting based on only 68 labeled architectures.

{
    \small
    \bibliographystyle{ieeenat_fullname}
    \bibliography{main}
}

\clearpage
\setcounter{page}{1}
\maketitlesupplementary

\section{Supplementary Materials}
\label{sec:supp}

We provide additional information and materials, e.g., predictor training, details of how ImageNet performance metrics are obtained, and further elaborate on our experimental setup, results and insights into Stable Diffusion U-Nets. 

\subsection{Predictor Training Setup}
\label{app:predictor}
We implement AutoBuild using PyTorch~\cite{NEURIPS2019_9015} and PyTorch-Geometric~\cite{Fey/Lenssen/2019}. We use a GATv2~\cite{brody2022how} GNN backbone with a hidden size of 32. There are 4 message-passing layers for PN/MBv3 and 6 message-passing layers for SDv1.4. Message-passing layers include a ReLU activation and residual skip-connections between hop layers. 

The FE-MLP applies a separate MLP to each node feature category to generate a single scalar per category. The only activation in the FE-MLP is an \texttt{abs} operation which is applied to the concatenation of each node feature scalar to produce the 0-hop embedding. Finally, graph embeddings at all hop-levels are generating using mean aggregation (Eq.~\ref{eq:aggr}). 

We train AutoBuild using Equation~\ref{eq:loss} (differentiable SRCC provided by \cite{blondel2020fast}) with a batch size of 128. We use a AdamW~\cite{loshchilov19AdamW} optimizer with an initial learning rate of $1e-4$ and weight decay of $1e-5$. Also, we standardize prediction labels using the sample mean and variance computed from the training data split. 

The number of epochs differs depending on task and the number of labeled samples we have. For the classification results in Sec.~\ref{sec:srcc}, we train for 200 epochs with $3k$ labeled architectures ($6k$ to generate Fig.~\ref{fig:mobilenet_onnx_ir}) at a training speed of 1 epoch/GPU second. For panoptic segmentation results in Sec.~\ref{sec:pan_seg_aiop}, we train for 300 epochs using the $1.3k$ `pseudo-labeled' MBv3 architectures from AIO-P~\cite{mills2023aiop}, while comparing against Pareto frontier generated from the 118 fully fine-tuned MBv3 architecture samples they provide. Finally, for SDv1.4, we have 68 samples and train each predictor (AutoBuild and Exhaustive Search) for 1000 epochs.

\subsection{Search Space Sizes}
\label{app:quant}

We provide formal mathematics to quantify the number of architecture modules present in each search space, as well as the size of each original macro-search space. Also, we calculate the size of the reduced search spaces used by \textbf{AutoBuild-NAS} and the \textbf{Layer-Insight}~\cite{mills2021profiling} from Sec.~\ref{sec:srcc}. 

\vspace{0.5em}
\noindent\textbf{Unique Architecture Modules.} 
PN and MBv3 both use MBConv blocks as layers. There are $9$ unique MBConv configurations formed as the cartesian product of kernel sizes $k \in \{3, 5, 7\}$ and expansion ratios $e \in \{3, 4, 6\}$. Stages 1-5 of PN and MBv3 contain 2-4 layers, for a total of $9^2+9^3+9^4= 81 + 729 +6561 = 7371$ unique 2-4 modules. This is sufficient for MBv3. 

PN has an additional 6th stage which always contain only a single layer, giving it $9$ additional subgraphs for $7380$ in total. However, in practice when enumerating over PN architecture modules and performing NAS, we merge the 5th and 6th stage for simplicity, meaning we consider 3-5 node subgraph modules of which there are $9^3+9^4+9^5 = 66339$. 

\vspace{0.5em}
\noindent\textbf{Mathematically Calculating Search Space Size.} 
Let $u$ represent a given stage, $l_u$ be the number of layers in stage $u$, and $\mathcal{B}_u$ be the set of unique layers available at stage $u$.  
Each stage is a subgraph that contains $l_u \in [l^{min}_u, l^{max}_u]$ layers, each sampled from $\mathcal{B}_u$ with replacement. The maximum number of unique stage subgraphs is therefore $|\mathcal{S}_u| = \sum_{l_u = l^{min}_u}^{l^{max}_u}|\mathcal{B}_u|^{l_u}$. Likewise, the total size of the search space is the cartesian product of all subgraphs across all stages, $\prod_{u = u^{min}}^{u^{max}}|\mathcal{S}_u|$.

For MBv3, these calculations work out to be $|\mathcal{S}_u| = \sum_{l_u=2}^{4} 9^{l_u}=7371$, for $u \in U = \{1, 2,.., 5\}$ as previously computed. The overall size is therefore $\prod_{1}^{5}7371 \approx 2.18\times10^{19}$. For PN, this number is multiplied by $9$ due to the additional 6th stage, for approximately $1.96\times10^{20}$ total architectures. It is infeasible if not intractable to consider all of these architectures, even using a predictor, yet by contrast the number of architecture module subgraphs we can consider is substantially smaller such that we can enumerate over them all. 

\begin{table*}[t]
    \centering
    \caption{Distribution statistics from 2.7k (90\% of 3k) random MBv3 architectures. $\mathcal{N}(\mu^h_m, \sigma^h_m)$ is the distribution of embedding norms for an accuracy predictor, while $\mathcal{N}(\mu^y_{u, l}, \sigma^y_{u, l})$ is the distribution of ImageNet top-1 accuracies [\%] per $(u, l)$. \textbf{Note}: $l = m+1$.}
    \scalebox{0.8}{
    \begin{tabular}{l|c|ccccc} \toprule
    & $\mathcal{N}(\mu^h_m, \sigma^h_m)$ & \multicolumn{5}{c}{$\mathcal{N}(\mu^y_{u, l}, \sigma^y_{u, l})$} \\ \midrule   
    $l$ & $m = l-1$ & $u=1$ & $u=2$ & $u=3$ & $u=4$ & $u=5$ \\ \midrule
    1 & $\mathcal{N}(2.24, 0.39)$   & -- & -- & -- & -- & --\\
    2 & $\mathcal{N}(13.83, 7.14)$  & $\mathcal{N}(76.77, 0.82)$ & $\mathcal{N}(76.77, 0.81)$ & $\mathcal{N}(76.69, 0.85)$ & $\mathcal{N}(76.42, 0.75)$ & $\mathcal{N}(76.61, 0.79)$\\
    3 & $\mathcal{N}(27.53, 8.67)$  & $\mathcal{N}(77.00, 0.79)$ & $\mathcal{N}(76.98, 0.78)$ & $\mathcal{N}(77.02, 0.76)$ & $\mathcal{N}(77.05, 0.70)$ & $\mathcal{N}(76.96, 0.76)$\\
    4 & $\mathcal{N}(40.94, 9.73)$ & $\mathcal{N}(77.08, 0.76)$ & $\mathcal{N}(77.12, 0.78)$ & $\mathcal{N}(77.13, 0.74)$ & $\mathcal{N}(77.41, 0.62)$ & $\mathcal{N}(77.26, 0.73)$\\
    \bottomrule
    \end{tabular}
    }
    \vspace{-3mm}
    \label{tab:dist_shift}
\end{table*}

\vspace{0.5em}
\noindent\textbf{Quantifying Reduced Search Spaces.} 
Also, the search space can be greatly reduced by limiting the size of $\mathcal{S}_u$, the number of possible subgraphs per stage. We reduce $|\mathcal{S}_u|$ by directly selecting a subset of architecture module subgraphs using AutoBuild in a data-driven manner. Specifically, for each target equation in the upper row of Fig.~\ref{fig:paretos}, we select the top-$K$ module subgraphs per unit where $K=25$. There are a maximum of $P=4$ target equations we consider at a time, for a total of $(4\times25)^5 =10^{10}$ architectures, still significantly less than the original search space by a factor of over $2\times10^9$. 

By contrast, \textbf{Layer-Insight} reduce $|\mathcal{B}_u|$ and $l_u^{max}$ by hand using human expert knowledge. For `MBv3-GPU', they restrict $l_u^{max}=3$ for stages 2, 4 and 5 and remove 2 MBConv blocks from consideration at all, resulting in a search space containing more than $4.6\times10^{14}$ architectures. Likewise, for `MBv3-CPU', they only constrain $l_u^{max}=3$ for stages 1, 2 and 3, which only reduces the search space to $2.9\times10^{16}$ architectures. Finally, `MBv3-NPU' removes $k=7$ from consideration so $|\mathcal{B}|_u=6$, giving a search space of size of ${\sim} 9\times10^{15}$. All of these are substantially larger than the reduced search spaces of AutoBuild. 

\subsection{Distribution Shift Statistics}
\label{app:dist}

We elaborate on how to calculate the statistics from Section 4.2, i.e., $\mathcal{N}(\mu^h_m, \sigma^h_m)$ and $\mathcal{N}(\mu^y_{u, l}, \sigma^y_{u, l})$. We also provide some empirical measurements to show the necessity of these calculations.

Recall that $\mathcal{N}(\mu^h_m, \sigma^h_m)$ is the distribution of node embedding norms learned by the GNN predictor, where $h$ refers to a hidden GNN embedding and $m$ parameterizes the hop-level. For each hop-level $m$, the statistics of this distribution are calculated after predictor training by a double for-loop that enumerates over every graph in the training set, and even node within each graph.  

By contrast, $\mathcal{N}(\mu^y_{u, l}, \sigma^y_{u, l})$ is the distribution of target labels $y_{\mathcal{G}}$ for graphs that have an architecture module subgraph with $l$ nodes positioned at stage $u$. These statistics are calculated by enumerating over the training dataset. Moreover, these statistics are specific to sequence-graphs and calculating them requires a sufficient number of graphs per $(u, l)$ tuple\footnote{We have around 1k for Sec.~\ref{sec:srcc} and around 430 for Sec. 5.2.}. 

We tabulate these statistics for an MBv3 accuracy predictor in Table~\ref{tab:dist_shift}. Note how the values of $\mu^h_m$ and $\sigma^h_m$ tend to grow monotonically with $m$. Also note how the accuracy distribution increases with $l$, yet differs for each stage $u$.

\subsection{Computation of ImageNet Metrics}
\label{app:imagenet}

We describe the evaluation protocol details for retrieving accuracy and latency of for all architecture AutoBuild constructs in the top row of Fig.~\ref{fig:paretos} or found by search in the bottom row of Fig.~\ref{fig:paretos}. Specifically, we obtain accuracy by splicing architecture subnets from the corresponding OFA~\cite{cai2020once} supernetwork and directly evaluating on ImageNet, which takes about 1 V100 GPU minute per architecture. 

We obtain latency measurements using predictors. These pure latency predictors are distinct from those mentioned in Figs.~\ref{fig:ofa_srcc} \& \ref{fig:mobilenet_onnx_ir} and the AutoBuild-NAS predictors described in Sec.~\ref{app:predictor} which we use to reduce the search space, but they are similar in design. Specifically, our latency predictors train on the OFA-MBv3/PN samples that \cite{mills2021profiling} provide\footnote{$50k$/$15k$/$5k$ unique architectures for the GPU/CPU/NPU devices with latency measurements for three input image sizes: \{$192^2$, $208^2$, $224^2$\}} for 500 epochs with a batch size of 32 to minimize the loss 

\begin{equation}
    \centering
    \label{eq:l1_predictor}
    \mathcal{L}_{Lat} = \texttt{MAE}(y_{\mathcal{G}}, y'_\mathcal{G}) + (1 - \rho(y_\mathcal{G}, y'_\mathcal{G})),
\end{equation}
where \texttt{MAE} is the Mean Absolute Error and $y_{\mathcal{G}}$ is the ground-truth latency of architecture $\mathcal{G}$. Also different from the AutoBuild predictors, the latency predictors use $k$-GNN~\cite{morris2019weisfeiler} message-passing layers with a hidden size of 128 and LeakyReLU activations. Finally, latency predictors compute graph embeddings using `sum' aggregation~\cite{you2020design} rather than the `mean' aggregation of Eq.~\ref{eq:aggr}. 

\subsection{NAS Evolutionary Algorithm Implementation}
\label{app:ea}

We use the random mutation-based EA provided by \cite{mills2021profiling}. This algorithm randomly samples an initial pool of architectures and establishes a population of the most Pareto-optimal architectures found. The algorithm then iteratively mutates architectures in the frontier by creating a predefined number of mutated architectures per iteration, and merging them into the existing Pareto frontier. The algorithm repeats for a set number of iterations. We calculate the total evaluation using $eval\_budget = initial\_archs + iter \times eval\_per\_iter$. We set a total evaluation budget of $250$ by setting $initial\_archs = 50$, $iter=4$ and $eval\_per\_iter=50$ for each experimental setup. 

By default, the algorithm performs NAS by mutating the layers of OFA-PN/MBv3 networks, e.g., adding, removing or altering the MBConv choice for a single layer per stage. Their code implements \textbf{Layer-Full} by default but also includes the search space restrictions required to run \textbf{Layer-Insight}, both of which we compare to in Fig. 5. We extend the code to provide support for \textbf{Stage-Full} and \textbf{AutoBuild-NAS}, both of which uses a different mutation mechanism where entire stages are changed at once rather than single layers. Note that both layer-wise and stage-wise mutation are equivalent in terms of evaluations. That is, whether a mutation changes one layer or one stage, the resultant architecture is still different from the original and must be evaluated separately to obtain a gauge performance.

\subsection{Additional FE-MLP Results}
\label{app:femlp}
We elaborate on why the operation type with the highest FE-MLP score in the right subfigure of Fig.~\ref{fig:mobilenet_onnx_ir} is `Add'. Further, we also provide additional FE-MLP results for PN and MBv3 that could not be included in the main manuscript due to space constraints.

\begin{figure}[t]
    \centering
    \includegraphics[width=3.25in]{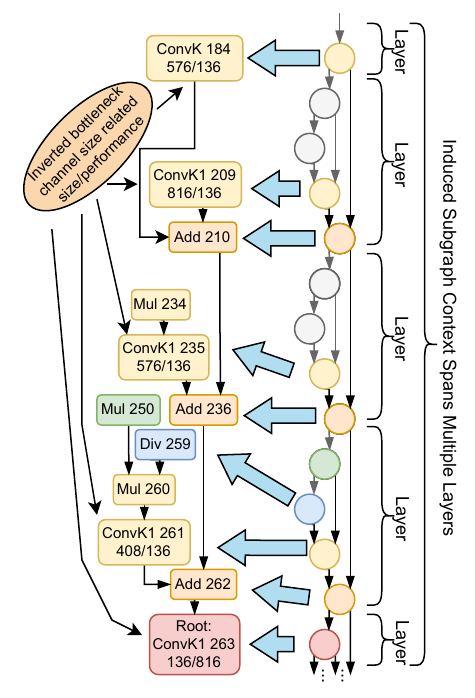}
    \caption{Annotated illustration of the 4-hop subgraph which achieved the highest score when training a AutoBuild accuracy predictor for PN/MBv3 using the `IR' format. Specifically, this subgraph comes from a large MBv3 architecture and is rooted at the first convolution in an MBConv block with $e = 6$. We annotate each node by operation type and topological sort index. Convolution nodes are further labeled with `Input Channel/Output Channel' numbers. Individual nodes are color-coded to match with the general area of the larger network they belong to.}
    \label{fig:mbv3_example}
    \vspace{-6mm}
\end{figure}

\vspace{0.5em}
\noindent\textbf{Importance of `Add'.} 
Both PN and MBv3 use long residuals to connect the start and end of MBConv blocks. These residual connections play a key role in network accuracy improvement and the add operator is often used to implement such skip-connections. As such, with sufficient hop-levels, subgraphs which incorporate add operations can span large sections of the overall network in order to accrue information from critical nodes that are far apart. For example, consider the MBv3 graph in Figure~\ref{fig:mbv3_example}. This graph contains several hundred nodes, yet the highest-ranked subgraph is rooted at `Conv 263'. This is a pointwise convolution\footnote{The depthwise convolutions with $k \in \{3, 5, 7\}$ are located in the middle of a block. The subgraph in Fig.~\ref{fig:mbv3_example} does not contain any of them.} which has input and output channels of $136$ and $816$, respectively, meaning that its channel expansion ratio is $e = \sfrac{816}{136} = 6$, the highest possible $e$ value in PN/MBv3. Since the input channel size is $136$, `Conv 263' is located in Stage 4 of MBv3, where changes in the number of layers/block choice can have a great impact on performance (see Fig. 5 of \cite{mills2021profiling}). Additionally, it is directly downstream from `Add 262', which receives input from `Conv 261' which has $e = 3$ and `Add 236'. `Add 236' is fed by `Conv 235' that has $e = 4$ and `Add 210', in turn fed by `Conv 209' ($e = 6$) and `Conv 184' ($e = 4$)\footnote{`Conv 184' is not preceded by an add node as it belongs to the first layer in the stage (`Conv 261'/`Add 262' is part of the 4th and final layer of the 4th MBv3 stage, while `Conv 263' is the first node in the 5th stage.), where downsampling occurs, and thus, no residual connection is included.}. 

This long chain of operation spans almost 80 node indices by topological sort, and the inclusion of `Div 259' signals that this subgraph is part of an MBv3 network, not PN. However, more importantly, the subgraph captures multiple pointwise convolutions which control the channel ratios in PN/MBv3. The number of channels directly affects the number parameters and FLOPs required to perform a forward pass - both of which are highly correlated with accuracy performance~\cite{mills2021profiling}. Of the 5 pointwise convolutions captured in the subgraph, only 1 of them has $e=3$, the worst option, while the remaining either have $e=4$ or $e=6$. Thus, not only does the subgraph belong to an MBv3 network, but the operations within it correspond to MBConv layers that perform heavy computations and likewise, help attain high performance. Since the formation of this subgraph fully depends on the residual `Add' operations, that operation type is assigned the highest FE-MLP score in Fig.~\ref{fig:mobilenet_onnx_ir}. 

\vspace{0.5em}
\noindent\textbf{FE-MLP Sequence Graph Scores.} 
The FE-MLP of AutoBuild provides a direct mean of measuring the scalar values for each choice of node feature, such as discrete kernel size, and interpreting these values as importance scores. Higher values correspond to higher target values through Eq.~\ref{eq:loss}. PN and MBv3 use MBConv blocks as layers, with a total of 9 types spanning  kernel sizes $k \in \{3, 5, 7\}$ and channel expansion ratios $e \in \{3, 4, 6\}$. We further enrich the node features by including the network input resolution and two position-based node features that encode a node stage $u$ and layer $l$ location. PN architectures contain 6 stages. Stages 1 to 5 contain 2 to 4 layers, while stage 6 always contains 1 layer. We perform an analysis of the feature importance for the PN and MBv3 macro-search spaces. Specifically, scores provided in the following figures stem from the same experiments performed to generate Fig.~\ref{fig:ofa_srcc} in the main manuscript.

\begin{figure}[t]
    \centering
    \subfloat{\includegraphics[width=1.64in]{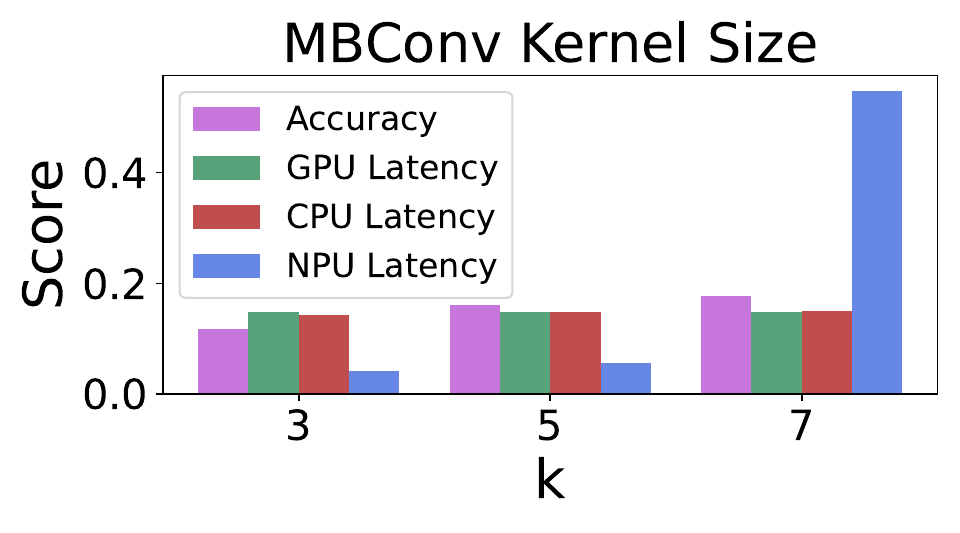}}
    \subfloat{\includegraphics[width=1.64in]{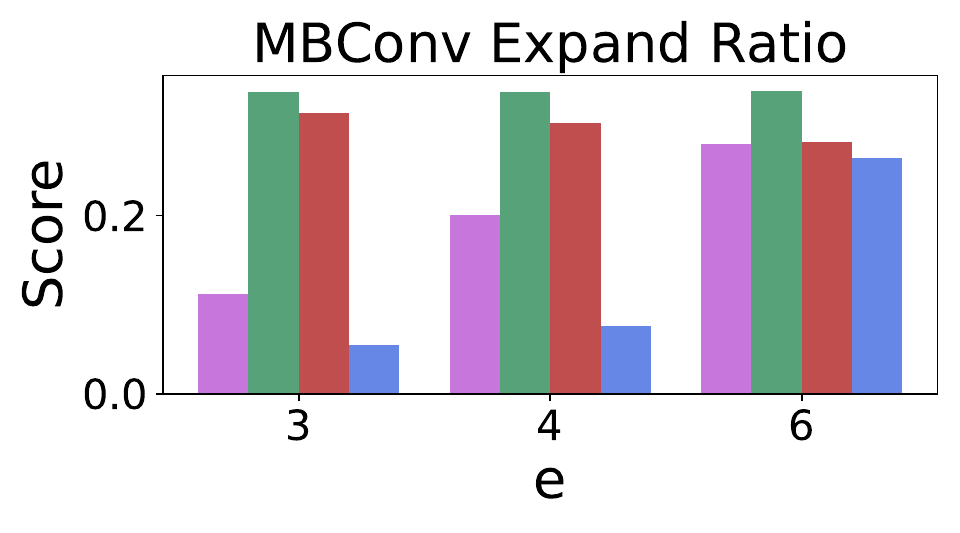}}
    \qquad
    \subfloat{\includegraphics[width=1.64in]{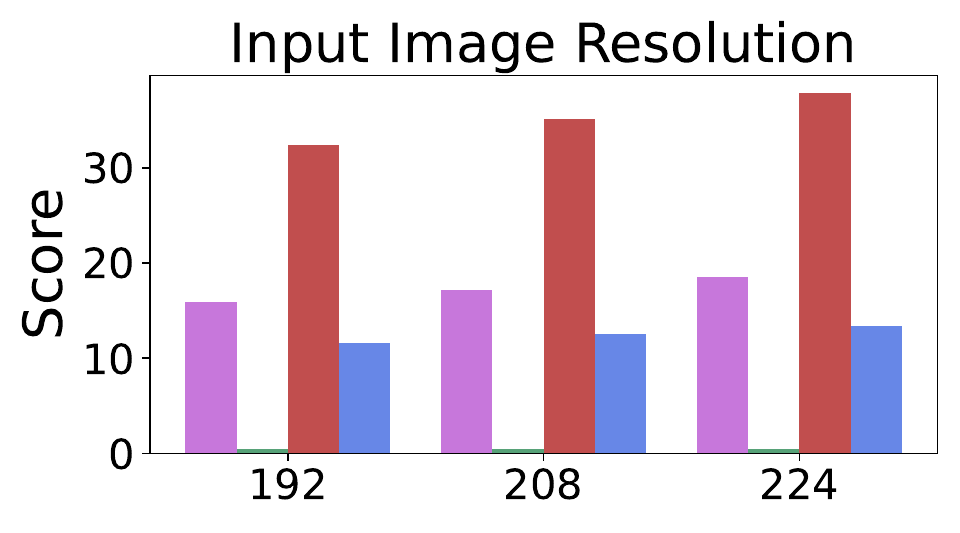}}
    \subfloat{\includegraphics[width=1.64in]{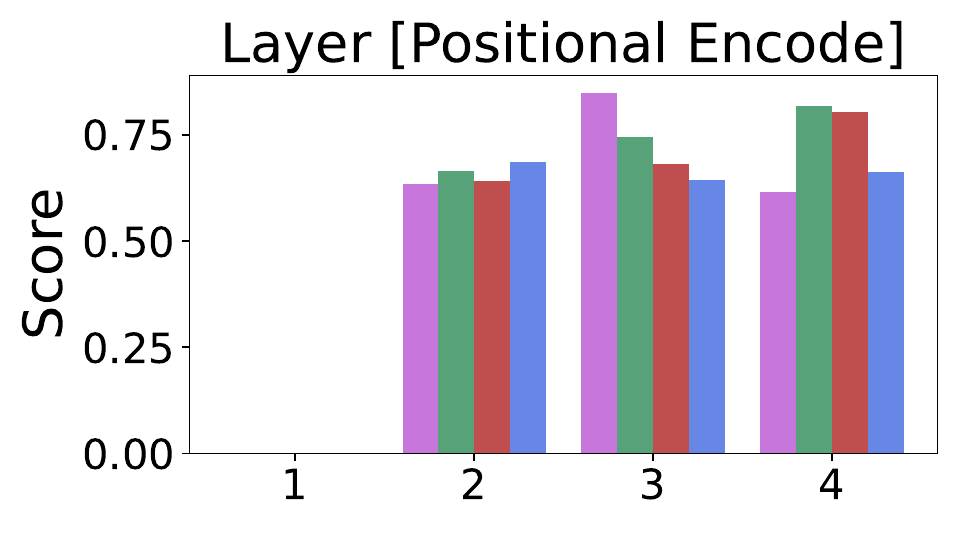}}
    \qquad
    \subfloat{\includegraphics[width=3.25in]{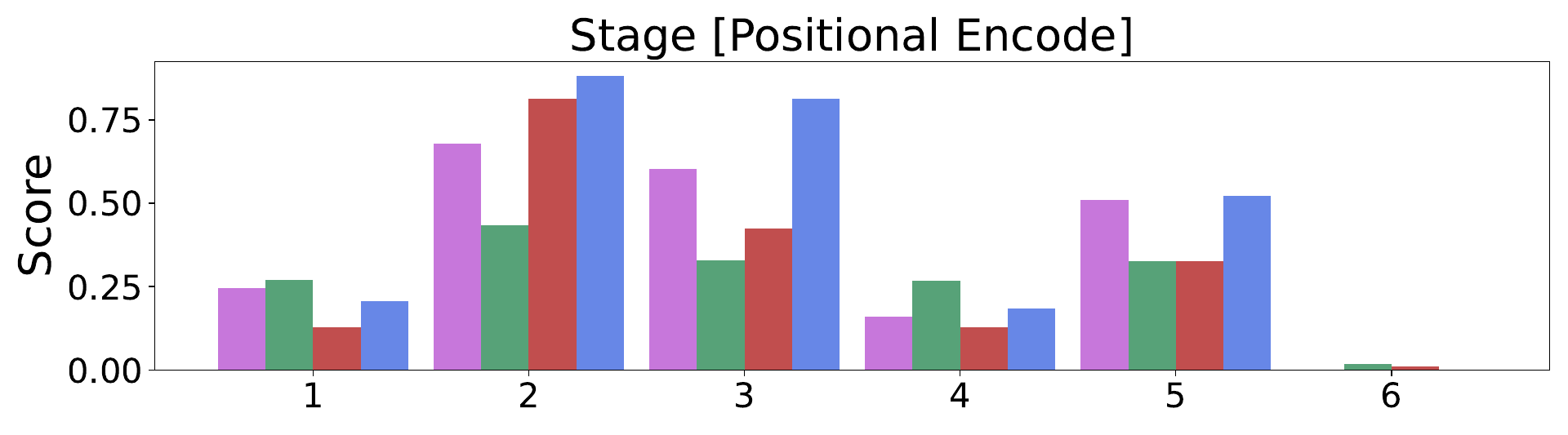}}
    \caption{FE-MLP 0-hop feature importance scores for PN. We train a AutoBuild predictor using the `custom' graph representation which provides 5 feature categories: Kernel, Expand, Resolution, Layers and Stages.}
    \label{fig:pn_feat_scores}
    \vspace{-4mm}
\end{figure}

\begin{figure}[t]
    \centering
    \subfloat{\includegraphics[width=1.65in]{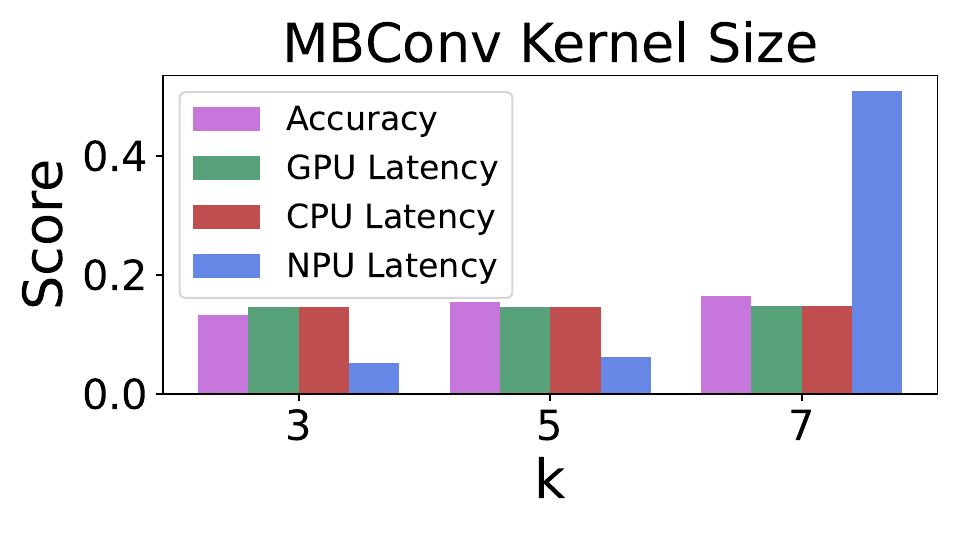}}
    \subfloat{\includegraphics[width=1.65in]{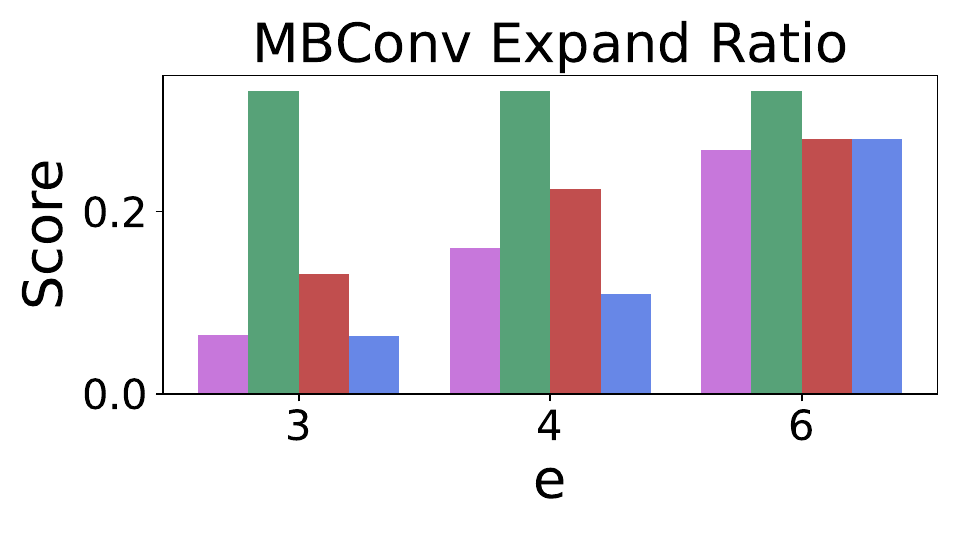}}
    \qquad
    \subfloat{\includegraphics[width=1.65in]{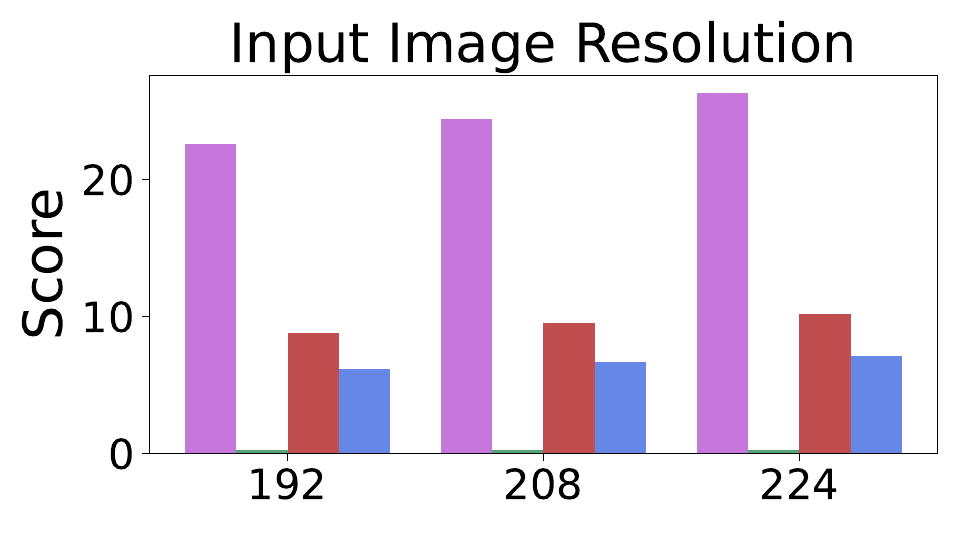}}
    \subfloat{\includegraphics[width=1.65in]{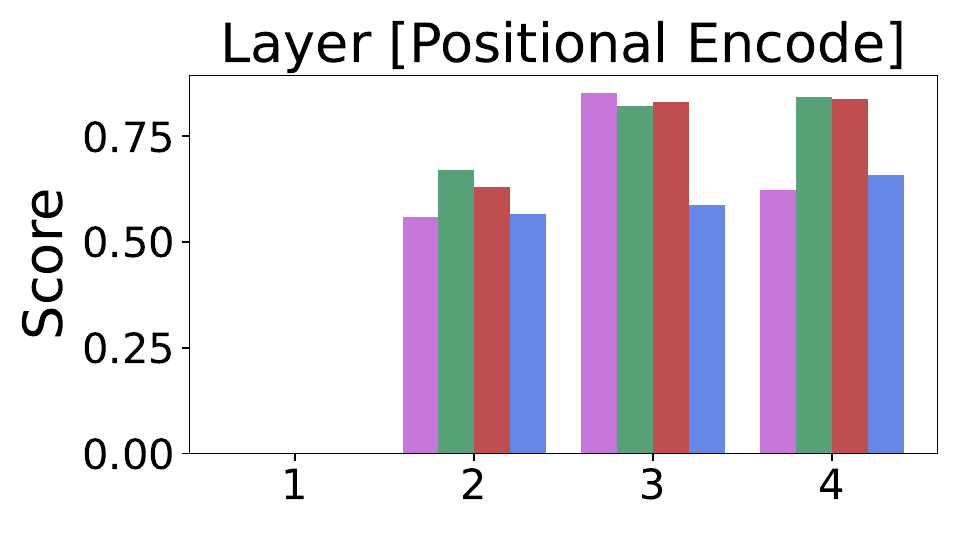}}
    \qquad
    \subfloat{\includegraphics[width=3.25in]{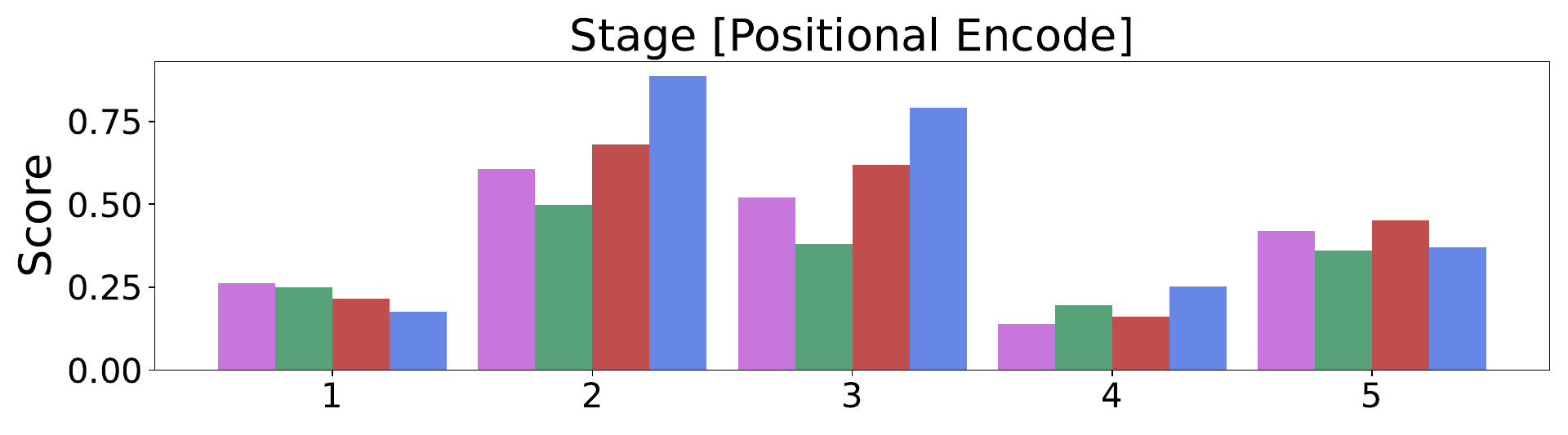}}
    \caption{FE-MLP 0-hop feature importance scores for MBv3. Same setup as Fig.~\ref{fig:pn_feat_scores}.}  
    \vspace{-4mm}
    \label{fig:mbv3_feat_scores}
\end{figure}

Figures~\ref{fig:pn_feat_scores} and \ref{fig:mbv3_feat_scores} illustrates the range of feature scores for the PN and MBv3 search spaces, respectively. Specifically, we consider accuracy, GPU latency, CPU latency and NPU latency metrics. We note that in both cases AutoBuild assigns disproportionate importance to kernel size 7 for NPU latency, which aligns with \cite{mills2021profiling} who found that $k=7$ disproportionately increases that metric. In other cases, kernel size, expansion ratio, and resolution generally follow linear trends or are mostly constant, e.g., expansion ratio $e$ for GPU latency on both search spaces, and PN CPU latency. Also, for PN, the resolution feature overall is essential when determining CPU latency, moderately important for accuracy and NPU latency, but has little to no effect in determining GPU latency. Yet, for MBv3, the resolution feature is emphasized more for accuracy than CPU latency.

In terms of positional features, we observe that a near-zero weight is universally assigned for $l=0$. Intuitively, this makes sense as the first layer always exists, while the 3rd and 4th layers are assigned higher values due to being optional. In~\cite{mills2021profiling}, architectures with stages containing 3 or more layers had higher GPU and CPU latency, especially if these layers were allocated to stages 2, 3 and 5, which obtained high scores across all metrics. Finally, for PN, stage $u=6$ is assigned a near-zero weight as it always exists and always contains 1 layer.

\subsection{Additional Panoptic Segmentation Results}
\label{app:pn}

Additionally, we also investigated the performance of AutoBuild on the ProxylessNAS macro-search space for Panoptic Segmentation. Much like MBv3, AIO-P~\cite{mills2023aiop} also provide sets of fully fine-tuned and `pseudo-labeled' architecture samples. As such, we apply AutoBuild to PN, using the same procedure detailed in Sec.~\ref{sec:pan_seg_aiop}. The only difference is we differ the target label equations as the PN PQ-FLOPs Pareto frontier differs from that of MBv3.

\begin{figure}
    \centering
    \includegraphics[height=1in]{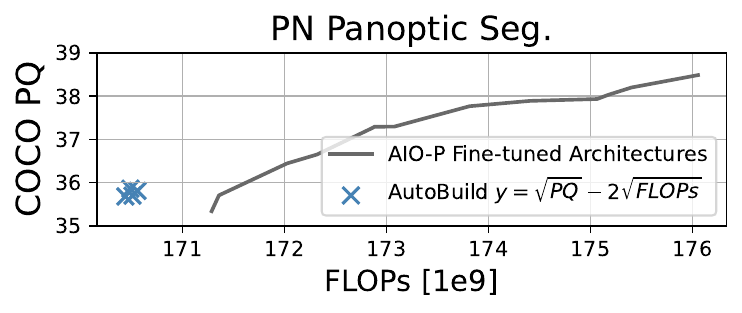}
    \caption{Results comparing AutoBuild PN architectures to the PQ-FLOPs Pareto frontier of fine-tuned architectures from \cite{mills2023aiop}.} 
    \label{fig:pn_pq}
    \vspace{-4mm}
\end{figure}

Figure~\ref{fig:pn_pq} illustrates our findings for a single target equation, $y=\sqrt{PQ} - 2\sqrt{FLOPs}$. While we are able to outperform the existing Pareto frontier, we originally crafted this target equation to focus on the center of the Pareto frontier. Specifically, we crafted the equation by considering three points on the Pareto frontier (Fig.~\ref{fig:pn_pq} grey line):
\begin{itemize}
    \item The `high' point: $(FLOPs, PQ) = (175.1, 37.9)$
    \item The `ideal' point: $(FLOPs, PQ) = (172.9, 37.4)$.
    \item The `low' point: $(FLOPs, PQ) = (171.3, 35.7)$.
\end{itemize}
The idea is to craft an equation where the `ideal' point receives a higher target than the `high' and `low' points. Specifically, using $y=\sqrt{PQ} - 2\sqrt{FLOPs}$, the predictor targets would be -20.31, -20.18 and -20.20 for the `high', `ideal' and `low' points respectively. While the `ideal' point has the highest target, it is closer to the `low' point than the `high' point, which influenced the architecture subgraph components found by AutoBuild. As such, and as stated in Sec.~\ref{sec:srcc}, improving on the target design scheme is definitely a direction for future work.

\subsection{Stable Diffusion 1.4 Inpainting Background}
\label{app:inpainting}
We provide additional details and experimental results regarding the Inpainting task described in Section~\ref{sec:inpainting} and elaborate on the AutoBuild macro-search space. Further, we provide illustrations of the best architectures found by AutoBuild and insights on the search space.

In the Inpainting task, we mask out certain parts of an image and rely on Stable Diffusion models to re-create the missing content. The baseline Inpainting model consists of two sub-modules: a VQ-GAN~\cite{esser2021taming} and a U-Net. 
The VQ-GAN is responsible for encoding/decoding images in the RGB domain to/from the latent domain.
The U-Net is responsible for removing noises from the input image.
The training of Diffusion models involves a forward diffusion process, where Gaussian noises are added to the input image in a step-wise fashion until the entire image becomes pure noise. And a reverse diffusion process, where the U-Net model attempts to remove the added noise and iteratively recovers the original image.
During inference, the U-Net generates new content from pure Gaussian noise.

\begin{figure}
    \centering
    \includegraphics[width=3in]{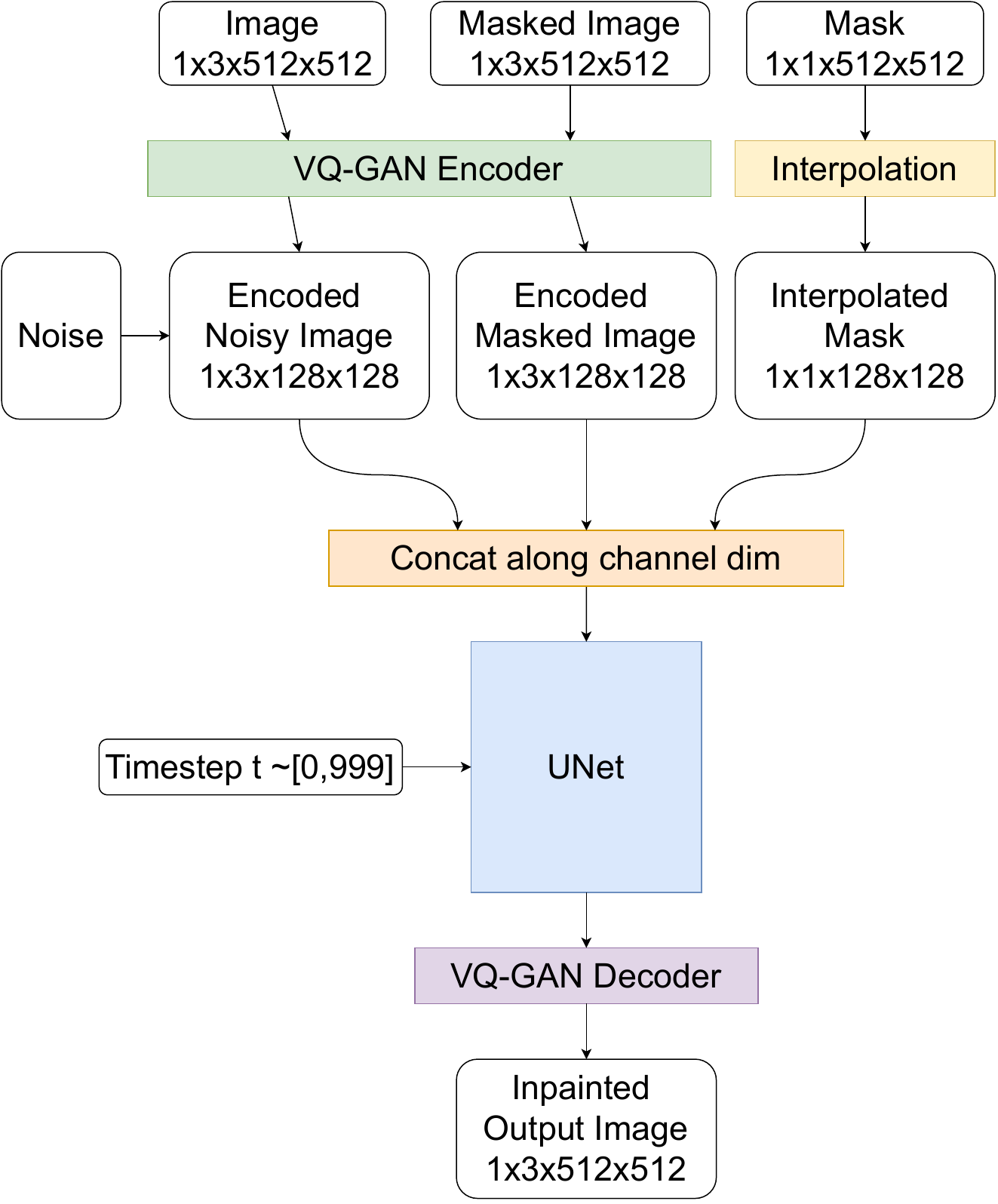}
    \caption{Overview of the Stable Diffusion 1.4 Inpainting model used in our experiments. Note that the U-Net model is our main focus for NAS-based optimization. The U-Net often need to run for multiple steps during inference.}
    \label{fig:inpainting_model}
    \vspace{-5mm}
\end{figure}

\begin{figure*}[t]
    \centering
    \subfloat{\includegraphics[width=6.8in]{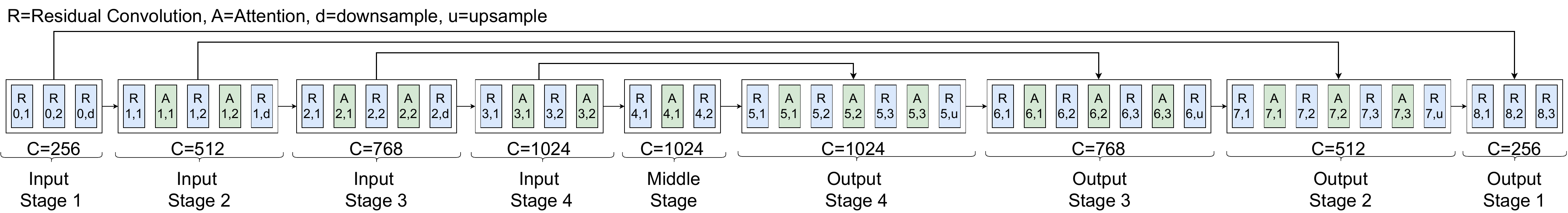}}
    \caption{Structure of the baseline U-Net model for Inpainting. E.g., `0,1' means the first block of the first stage.} 
    \label{fig:U-Net_model}
    \vspace{-3mm}
\end{figure*}

In our experiments we use the open-source Stable Diffusion v1.4~\cite{Rombach_2022_CVPR} VQ-GAN and U-Net model, but we only optimize the U-Net architecture using NAS.
To train the U-Net model, the original image, the masked image and the mask are first encoded into latent space by either the VQ-GAN encoder or interpolation. 
Next, random Gaussian noise is added, and the U-Net is trained to recover the noise-free image in the latent space.
An overview of our Inpainting model is illustrated in Figure~\ref{fig:inpainting_model}.
The U-Net model can be further dissected into input, middle and output branches. 
The input and output branches contains multiple stages with residual connections in-between (e.g., the output of stage 2 from the input branch will be added to the output of stage 2 from the output branch).
Each stage further partitions into Residual Convolution and Attention Blocks. 
At the end of input stages 1, 2 and 3, there are downsampling blocks. Conversely, at the end of output stages 2, 3 and 4, there are upsampling blocks. The downsampling/upsampling blocks adjust the latent tensor height, width and channels (HWC). Figure~\ref{fig:U-Net_model} provides a compact view of the SDv1.4 U-Net. 

\subsection{AutoBuild for FID Minimization}
\label{app:autobuild_fid}
We consider two distinct SDv1.4-based search spaces and fine-tuning regimes. The first of which pertains to all results listed and illustrated in Sec.~\ref{sec:inpainting} and is further detailed in this subsection. Here, the objective is to minimize the FID. The second search space aims to find Pareto-optimal architectures where a secondary hardware metric, e.g., latency, is taken into account. All results and discussion pertaining to the second search space can be found in Sec.~\ref{app:hw_aware_sdv14}.

\begin{figure*}[t]
    \centering
    \subfloat{\includegraphics[width=6.8in]{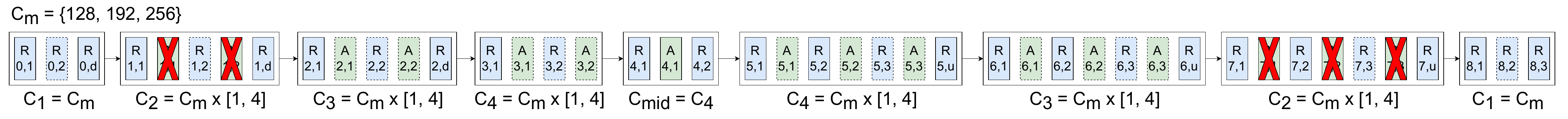}}
    \caption{Illustration of the searchable attributes in the SDv1.4 U-Net. Long residual connections are removed from this drawing but still present in all candidate architectures. The attention blocks in Stage 2 are always removed. We do not alter the middle stage but note that its channels must match that of the stage 4 input/output modules. 
    We allow for a global channel value $C_m \in \{128, 192, 256\}$. The number of channels in stages 2-4 is equal to $C_m$ times a multiplier in $[1, 2, 3, 4]$, while the channels of Stage 1 are always $C_m$. We can also remove attention blocks from the input/output stages 3 and 4 independently of each other, as well as remove at least one residual convolution block per input/output stage. Blocks with dotted borders denote optional layers, while blocks bound by solid lines are mandatory.} 
    \label{fig:U-Net_search}
    \vspace{-3mm}
\end{figure*}

\vspace{0.5em}
\noindent\textbf{FID Minimization Search Space \& Evaluation.} 
Our search space is designed around the SDv1.4 U-Net. We always remove the attention blocks from Stage 2, and then configure the rest of the U-Net according to several searchable attributes, given below:

\begin{itemize}
  \item \textbf{U-Net global channel size $C_m$}: Baseline has 256 channels, our searchable candidates are [128, 192, 256].
  \item \textbf{Stage channel multipliers}: Baseline has multipliers [1, 2, 3, 4] corresponding to channels 256, 512, 768, 1024 for stage 1, 2, 3, 4 of the input/output branches. We add 7 other combinations as searchable candidates (e.g., [1, 1, 2, 4], [1, 3, 4, 2]). 
  \item \textbf{Number of residual blocks}: Baseline has 2 residual blocks per stage for the input branch and 3 for the output branch. We allow the optional removal of 1 residual block for any number of stages.
  \item \textbf{Number of attention blocks}: We allow the optional removal of all attention blocks in stages 3 and/or 4.
\end{itemize}
An illustration of the search space choices is provided by Figure~\ref{fig:U-Net_search}. 
To represent candidate architectures as sequence graphs, we encode the following node features:
\begin{itemize}
    \item \textbf{Operation type}, which is one of: `Residual Convolution', `Attention', `Downsample' or `Upsample'.
    \item \textbf{Channel size}. \textbf{Note:} To instantiate an SDv1.4 architecture variant as a functional neural network, the global channel size $C_m$ must be consistent across all nodes. Also, the channel multiplier values for nodes in an input stage and its corresponding output stage must be the same, e.g., if the multiplier for Input Stage 2 is $3$, the multiplier for Output Stage 2 must also be $3$. 
    \item \textbf{Positional embedding} describing the which section (input, middle, or output) and stage a node is located in. 
\end{itemize}

After combining the searchable parameters and applying the constraints on $C_m$ and stage-channel multipliers, we are left with a search space containing approximately 800\textit{k} U-Net candidates. To evaluate a candidate, we first inherit weights from the original SDv1.4, then fine-tune for 20\textit{k} steps on the Places365~\cite{zhou2017places} training set, which contains 1.7 million images. We use 2 Nvidia Tesla V100 GPUs with 32GB of VRAM each to fine-tune, with a batch size of 7 per GPU. Then we compute its FID on a held-out Places365 validation subset containing 3k images where the Inpainting masks are predefined and consistent across different architecture evaluations. The whole process can take between 1-2 days depending on candidate architecture.

\vspace{0.5em}
\noindent\textbf{Prediction with ${<}100$ Samples.} 
We provide additional details on how we use AutoBuild to select high-performance architectures. Specifically, each predictor in the ensemble trains for 1000 epochs with batch gradient descent, e.g., the batch size is equal to the size of the training set. The GNN predictor contains 6 message passing layers to accommodate output stages 2, 3 and 4 which can span up to 7 nodes. Since we only have 68 samples and the number of hops can vary between 0 and 6, we do not have enough data to compute the statistics required for Eq.~\ref{eq:distshift}. Instead, we normalize subgraph scores to a normal distribution $\mathcal{N}(0, 1)$, then take the top-$5$ subgraphs per input/output stage\footnote{We ignore the Middle Stage from consideration and simply select the channel figuration that will match with the Stage 4 Input/Output subgraphs.}, dividing $K=5$ between different subgraph sizes. For example, when selecting subgraphs for Stage 1, which can have 1-2 hops, we will select two 1-hop subgraphs and three 2-hop subgraphs, while for Output Stages 2, 3 and 4, which can span 2-6 hops, we select the best subgraph for each hop-size. 

\begin{figure*}[t]
    \centering
    \subfloat{\includegraphics[width=6.8in]{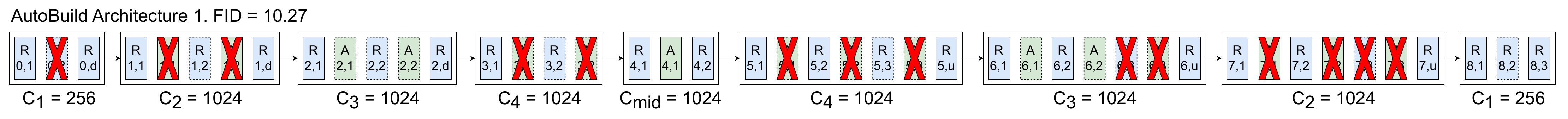}}
    \qquad
    \subfloat{\includegraphics[width=6.8in]{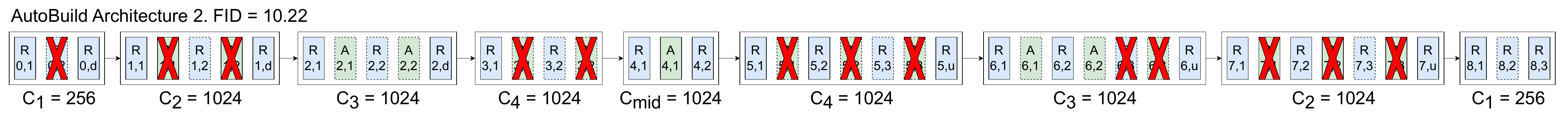}}
    \qquad
    \subfloat{\includegraphics[width=6.8in]{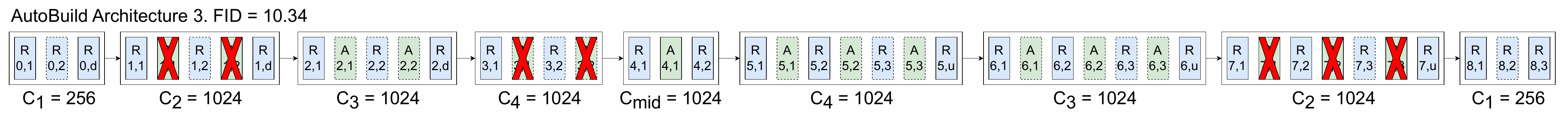}}
    \qquad
    \subfloat{\includegraphics[width=6.8in]{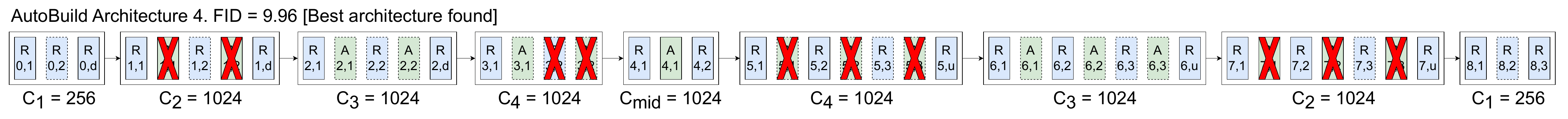}}
    \caption{Architectures found by AutoBuild, annotated with FID performance.} 
    \label{fig:autobuild_sdv14}
\end{figure*}

\begin{figure*}[t]
    \centering
    \subfloat{\includegraphics[width=6.8in]{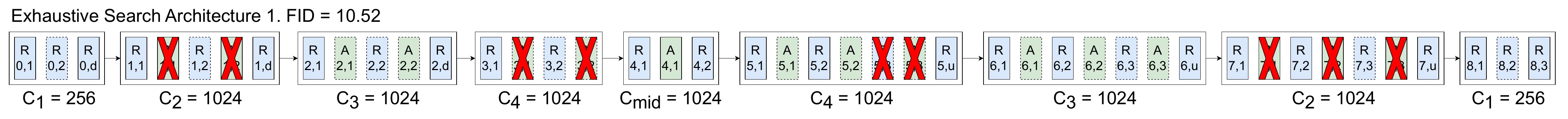}}
    \qquad
    \subfloat{\includegraphics[width=6.8in]{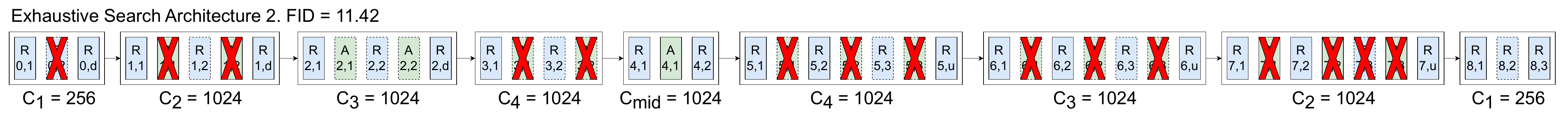}}
    \qquad
    \subfloat{\includegraphics[width=6.8in]{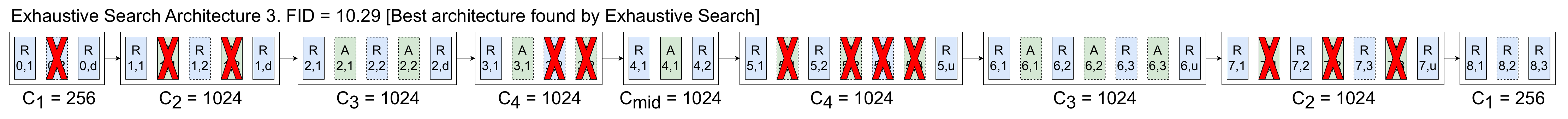}}
    \qquad
    \subfloat{\includegraphics[width=6.8in]{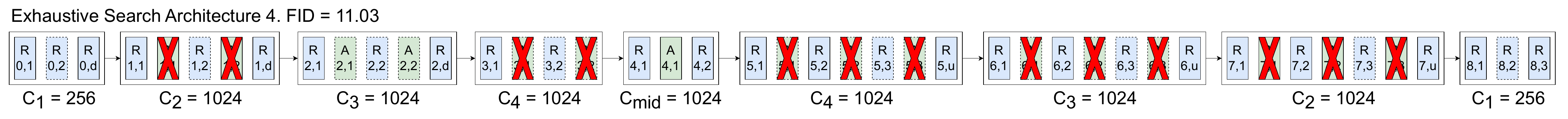}}
    \caption{Architectures found by Exhaustive Search, annotated with FID performance.} 
    \label{fig:exhaust_sdv14}
\end{figure*}

The top-4 architectures are then constructed by finding the top-$4$ subgraph combinations whose sum of scores is the highest, provided they abide by the aformentioned channel constraints to ensure network functionality. Figures~\ref{fig:autobuild_sdv14} and \ref{fig:exhaust_sdv14} illustrate the top-$4$ architectures found by AutoBuild and Exhaustive Search, respectively, annotated with FID.

\begin{table}[t]
    \centering
    \caption{Distribution and range of feature categories for SDv1.4 generated by the AutoBuild predictor that attained the highest FE-MLP 0-hop SRCC.}
    \scalebox{0.8}{
    \begin{tabular}{lcc} \toprule
    \textbf{Feature} & \textbf{Distribution} & \textbf{Range} \\ \midrule
    Operation & $\mathcal{N}(0.53, 0.21)$ & $[0.00, 0.77]$ \\
    Channels & $\mathcal{N}(0.48, 0.04)$ & $[0.38, 0.53]$ \\
    Section Idx. & $\mathcal{N}(0.52, 0.60)$ & $[0.02, 1.30]$ \\
    Stage Idx. & $\mathcal{N}(0.43, 0.22)$ & $[0.26, 0.79]$ \\
    \bottomrule
    \end{tabular}
    }
    \vspace{-5mm}
    \label{tab:sdv14_femlp}
\end{table}

\vspace{0.5em}
\noindent\textbf{SDv1.4 Search Space Insights.} 
We provide some additional search space insights from the FE-MLP of the AutoBuild predictor which achieved the highest 0-hop SRCC ($0.6464$). Table~\ref{tab:sdv14_femlp} quantifies the distribution and range of importance scores for each feature category. First, we note that the feature which attains the highest value is Section Index. Specifically, input stages are assigned a high score of $1.3$, output stages are assigned $0.15$ while the Middle Stage is assigned a low value of $0.08$, notable as the best architecture found (AutoBuild Architecture 4) allows attention in every input stage it can. In terms of Stage Index, importances are given in descending order as Stage $4=0.79$, Stage $2=0.41$, Stage $3=0.26$ and Stage $1=0.26$ which emphasizes the input/output stages closest to the middle, as well as Stage 2, where attention is part of the original model but not permitted in our search space.

\begin{figure}[t]
    \centering
    \includegraphics[width=2in]{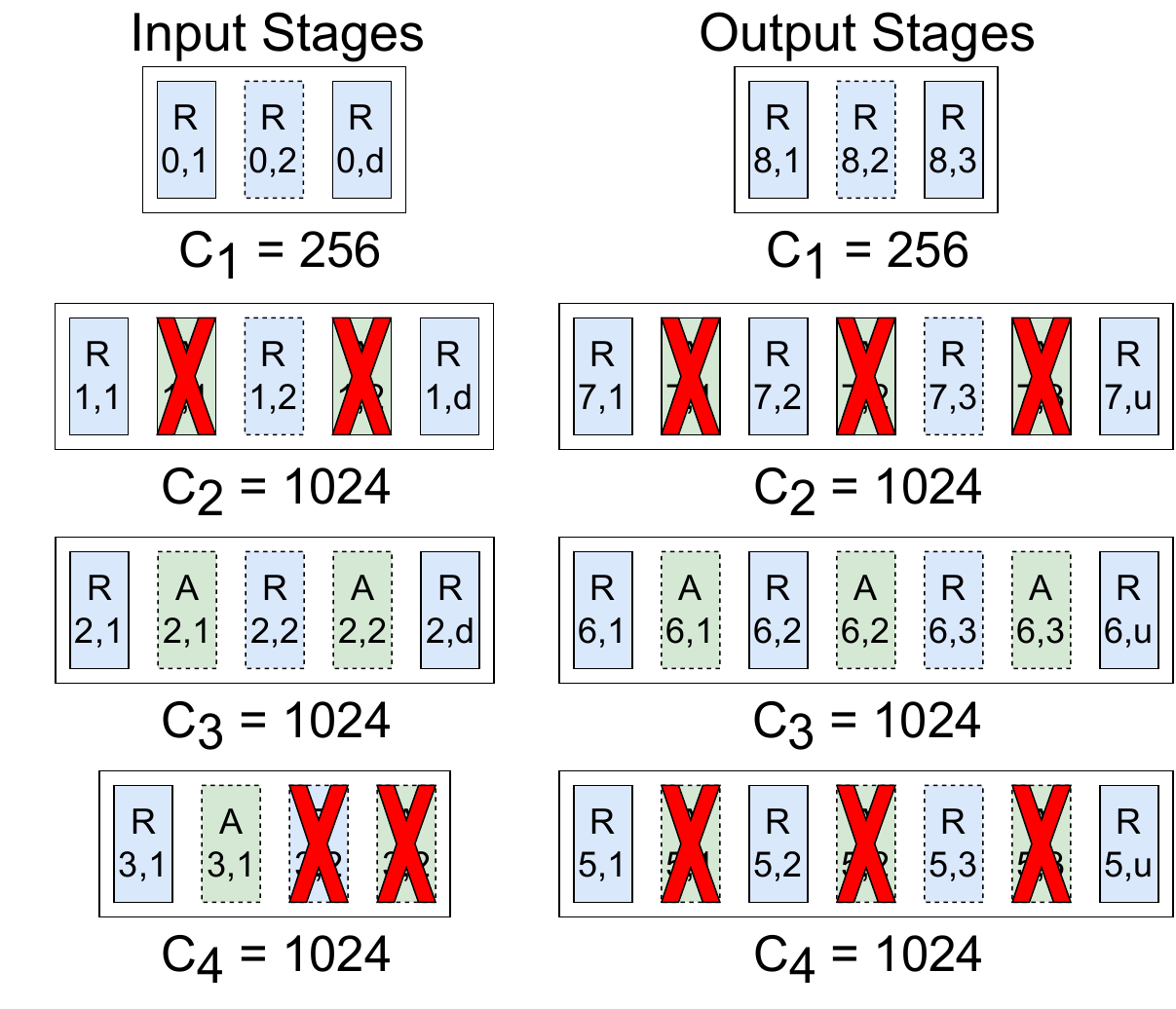}
    \caption{Best AutoBuild SDv1.4 input/output subgraphs.}
    \label{fig:sdv14_best_sgs}
    \vspace{-3mm}
\end{figure}

Operation importance scores are as follows, in descending order: Upsample $=0.77$, Residual Convolution $=0.63$, Attention $=0.33$ and Downsample $=0.00$. This may be a way of compensating for the high score assigned to nodes in the input stages while not placing importance on Output Stage 1, as it is the only output stage that lacks an upsampling operation. Finally, channel size is correlated with importance score: more channels is better, as evidenced by all architectures in Figs.~\ref{fig:autobuild_sdv14} and \ref{fig:exhaust_sdv14} setting $C_m = 256$ and using the highest-possible multipliers in stages 2, 3 and 4.  Finally, Figure~\ref{fig:sdv14_best_sgs} illustrates the best input and output architecture module subgraphs found by AutoBuild for our SDv1.4 macro-search space.

\subsection{AutoBuild for Pareto-optimal U-Nets}
\label{app:hw_aware_sdv14}

In addition to the search space and AutoBuild results present in Sections~\ref{sec:inpainting} and \ref{app:autobuild_fid}, we designed a second search space and training regime. The objective of this search space and experiments was to find smaller SDv1.4 U-Net variants that achieved Pareto-optimal performance by minimizing the FID and a hardware-friendliness metric, e.g., GPU latency.

\vspace{0.5em}
\noindent\textbf{Hardware-aware Search Space \& Evaluation.} 
The second U-Net search space is a subset of the first, only containing approximately $100k$ architectures compared to the $800k$ in the original. We now iterate the changes made to differentiate the second search space: 

\begin{itemize}
    \item \textbf{U-Net global channel size $C_m$}: Fixed to always be 256.
    \item \textbf{Stage channel multipliers}: We fix these so that the number of channels is never more than the original SDv1.4 U-Net, e.g., Stage 4 can choose from multipliers in $[1, 2, 3, 4]$, but Stage 2 can only choose from $[1, 2]$.
    \item \textbf{Number of residual/attention blocks}: Unchanged.
\end{itemize}

These changes produce a new search space where the original SDv1.4-U-Net is the largest architecture. In terms of fine-tuning and evaluation, we still use the Places365 dataset. Specifically, we use the same $3k$ masked images from the validation set to generate FID scores as before. However, FID for architectures in the first and second search spaces are not strictly comparable as we increased the number of training steps to $30k$, up from $20k$ before.

We fine-tune and evaluate $90$ random architectures from the second search space. In addition to measuring the FID score, we also consider three hardware metrics: U-Net inference latency in an Nvidia V100 GPU, the number of Floating Point Operations (FLOPs) required for a forward pass, and the number of U-Net parameters.

\vspace{0.5em}
\noindent\textbf{Evaluation and Results.} 
Our experimental procedure is largely unchanged from Sections~\ref{sec:inpainting} and \ref{app:autobuild_fid} in that we still consider an \textbf{Exhaustive Search} baseline method. However, this time we evaluate two variants of AutoBuild on the second search space, \textbf{AutoBuild Unconstrained} and \textbf{AutoBuild Constrained}. We differentiate these methods by how they account for module subgraphs with different sizes when constructing high-performance architectures: 

\begin{itemize}
    \item \textbf{AutoBuild Constrained} follows the procedure from Sec.~\ref{app:autobuild_fid}, ensuring that a variety module subgraphs with different hop sizes are selected when composing the top-$K$ set for a given stage.
    \item \textbf{AutoBuild Unconstrained} does not follow this procedure. Instead, we just normalize all subgraph scores for a given stage into $\mathcal{N}(0, 1)$ and then take the top-$5$ subgraphs per stage. 
\end{itemize}

\begin{figure}[t]
    \centering
    \subfloat[$y=-FID - Latency$]{\includegraphics[width=2.85in]{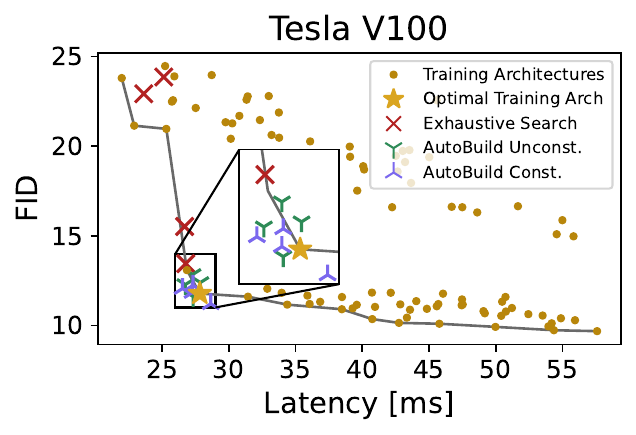}}  
    \qquad
    \subfloat[$y=-FID-\dfrac{Params}{200}$]{\includegraphics[width=2.85in]{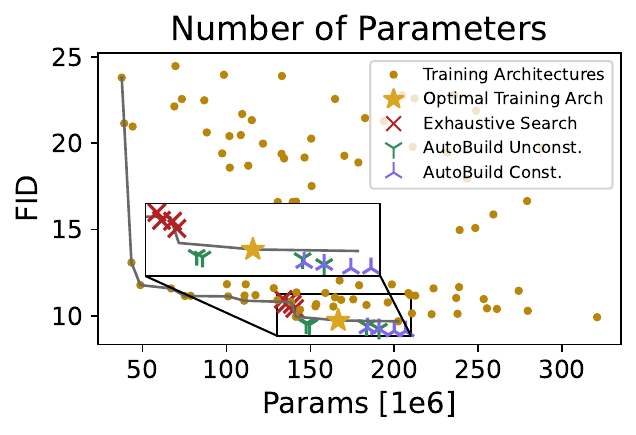}}  
    \qquad
    \subfloat[$y=-FID-\dfrac{FLOPs}{400}$]{\includegraphics[width=2.85in]{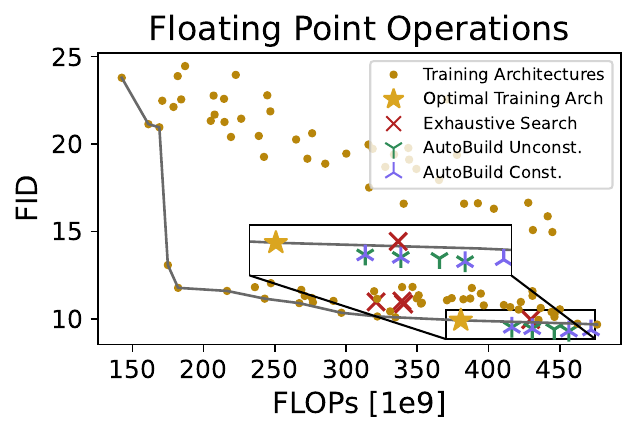}}   
    \caption{FID vs. Hardware-friendliness Pareto frontiers (lower values are better for both) for the second SDv1.4 search space (Sec.~\ref{app:hw_aware_sdv14}). Subfigure captions refer to the target equations that AutoBuild and Exhaustive Search aim attempt to maximize. `Training Architectures' refers to the 90 random fine-tuned U-Nets, while the `Optimal Training Arch' gold star indicates the training architecture that maximizes the target equation. }
    \label{fig:sdv14_constrained}
\end{figure}

Figure~\ref{fig:sdv14_constrained} illustrates the resultant Pareto frontiers for a trio of hardware metrics. Note how in every case AutoBuild is able to consistently find better U-Net architectures that push beyond the original Pareto frontier formed by the 90 random training architectures. In addition, the architectures found by AutoBuild are generally much closer to the `Optimal Training Arch', which is the training architecture that maximizes the target objective, than those found by Exhaustive Search.

When comparing AutoBuild Constrained and AutoBuild Unconstrained, we find that the `Constrained' version can more consistently construct architectures that surpass the original Pareto frontier, e.g., on the V100 GPU. A downside of this approach is that architectures found by AutoBuild Constrained actually deviant further away from the Optimal Training Architecture than those constructed by AutoBuild Unconstrained, e.g., in the FLOPs and parameter frontiers. Finally, also common to the FLOPs and parameter Pareto frontiers, is not uncommon for both of these methods to construct the same architectures.

\end{document}